 \documentclass[final,5p,times,twocolumn]{elsarticle}

\usepackage{amssymb, amsmath, float, stfloats, multirow, subfigure, booktabs, arydshln, xcolor, pifont, makecell, algorithm, algorithmic, threeparttable, placeins}
\usepackage{xcolor}
\usepackage{colortbl}

\definecolor{e}{RGB}{73,157,191}
\usepackage[colorlinks,
           linkcolor=e,
           anchorcolor=e,
           citecolor=e]{hyperref}
\usepackage[figuresright]{rotating}

\begin{document}
\begin{frontmatter}
\title{Dynamically Allocated Interval-Based Generative Linguistic Steganography with Roulette Wheel}
\tnotetext[t1]{\textbf{The paper has been accepted by \textit{Applied Soft Computing}.}}
\tnotetext[t1]{\textit{\textbf{Thanks for the support provided by MindSpore Community.}}}
\tnotetext[t2]{This work is supported by the National Natural Science Foundation of China (Grant U21B2020) and supported by BUPT Excellent Ph.D. Students Foundation (Grant CX2023120, CX20241055).}

\author{Yihao Wang}
\ead{yh-wang@bupt.edu.cn}

\author{Ruiqi Song}
\ead{songrq123@bupt.edu.cn}

\author{Lingxiao Li}
\ead{Lingxiao-Li@bupt.edu.cn}

\author{Ru Zhang\textsuperscript{*}}
\ead{zhangru@bupt.edu.cn}

\author{Jianyi Liu}
\ead{liujy@bupt.edu.cn}

\cortext[cor1]{Corresponding author}
\address{School of Cyberspace Security, Beijing University of Posts and Telecommunications, Beijing, China.}

\begin{abstract}
Existing linguistic steganography schemes often overlook the conditional probability (CP) of tokens in the candidate pool, allocating the one coding to all tokens, which results in identical selection likelihoods. This approach leads to the selection of low-CP tokens, degrading the quality of stegos and making them more detectable. This paper proposes a scheme based on the interval allocated, called DAIRstega. DAIRstega first uses a portion of the read secret to build the roulette area. Then, this scheme uses the idea of the roulette wheel and takes the CPs of tokens as the main basis for allocating the roulette area (i.e., the interval length). Thus, tokens with larger CPs are allocated more area. The secret will have an increased likelihood of selecting a token with a higher CP. During allocation, we designed some allocation functions and three constraints to optimize the process. Additionally, DAIRstega supports prompt-based controllable generation of stegos. Rich experiments show that the proposed embedding way and DAIRstega perform better than the existing ways and baselines, which shows strong perceptual, statistical, and semantic concealment, as well as anti-steganalysis ability. It can also generate high-quality longer stegos, addressing the deficiencies in this task. DAIRstega is confirmed to have potential as a secure watermarking, offering insights for its development. Our codes and data are available at: \href{https://github.com/WangYH-BUPT/DAIRstega}{https://github.com/WangYH-BUPT/DAIRstega}.
\end{abstract}

\end{frontmatter}


\section{Introduction}\label{sec1}
Steganography, a technology of information hiding \cite{Shannon}, embeds secrets within media such as texts \cite{Yang2023TDSC, VAE-Stega} and images, obtaining steganographic media that is perceptually indistinguishable from normal media. Only authorized receivers can perceive whether the media is steganographic and accurately extract the secrets, thereby protecting privacy. Benefiting from the ability of text to be transmitted losslessly over public channels, research on linguistic steganography has explosive growth \cite{Lu2023IJCNN, Hi-Stega, Yang2023TDSC}. To increase the anti-detection of steganographic text (\underline{stego}), the focus of work has shifted from modified schemes \cite{Huo2016ICCC} represented by synonym replacement \cite{Kim2010} to generative schemes \cite{Topic, VAE-Stega}.

The concealment of stego determines the success of covert communication to a great extent. Depending on the constraints, the concealment is manifested in three aspects: perceptual, statistical, and semantic. ``Perceptual concealment'' focuses on ensuring stegos with complement and fluency. ``Statistical concealment'' requires the distribution of stegos to be close to that of the natural texts (\underline{covers}). ``Semantic concealment'' aims to stego that is coherent and semantically controllable. The concealment of stego mainly depends on the design of the steganography embedding way. An excellent embedding way ensures that the conditional probability (CP) distribution of the text is not changed as much as possible while embedding the secret. In this way, stegos are difficult to perceive and detect by unauthorized receivers. Most existing schemes have made efforts in this regard. Yang et al. \cite{RNN-Stega} and Ding et al. \cite{Ding2023TCDS} encoded each token in the candidate pool with Fixed length coding (FLC) and only used secret information to determine the next token to achieve the purpose of embedding. Yang et al. \cite{VAE-Stega} and Zhou et al. \cite{GAN-APD} designed the Huffman coding-based (HC) embedding, which generates stegos with good completeness and fluency, improving perceptual concealment. Zhang et al. \cite{ADG} proposed the Adaptive dynamic grouping coding-based (ADG) embedding. They achieved strong statistical concealment. Kaptchuk et al. \cite{Meteor} and Wang et al. \cite{Hi-Stega} proposed the Arithmetic coding-based (AC) embedding. Lu et al. \cite{Lu2023IJCNN} used contextual learning and variable length coding-based (VLC) to generate stegos. It improves semantic concealment.

\begin{figure}[!htbp]
	\centering
	\includegraphics[width=0.48\textwidth]{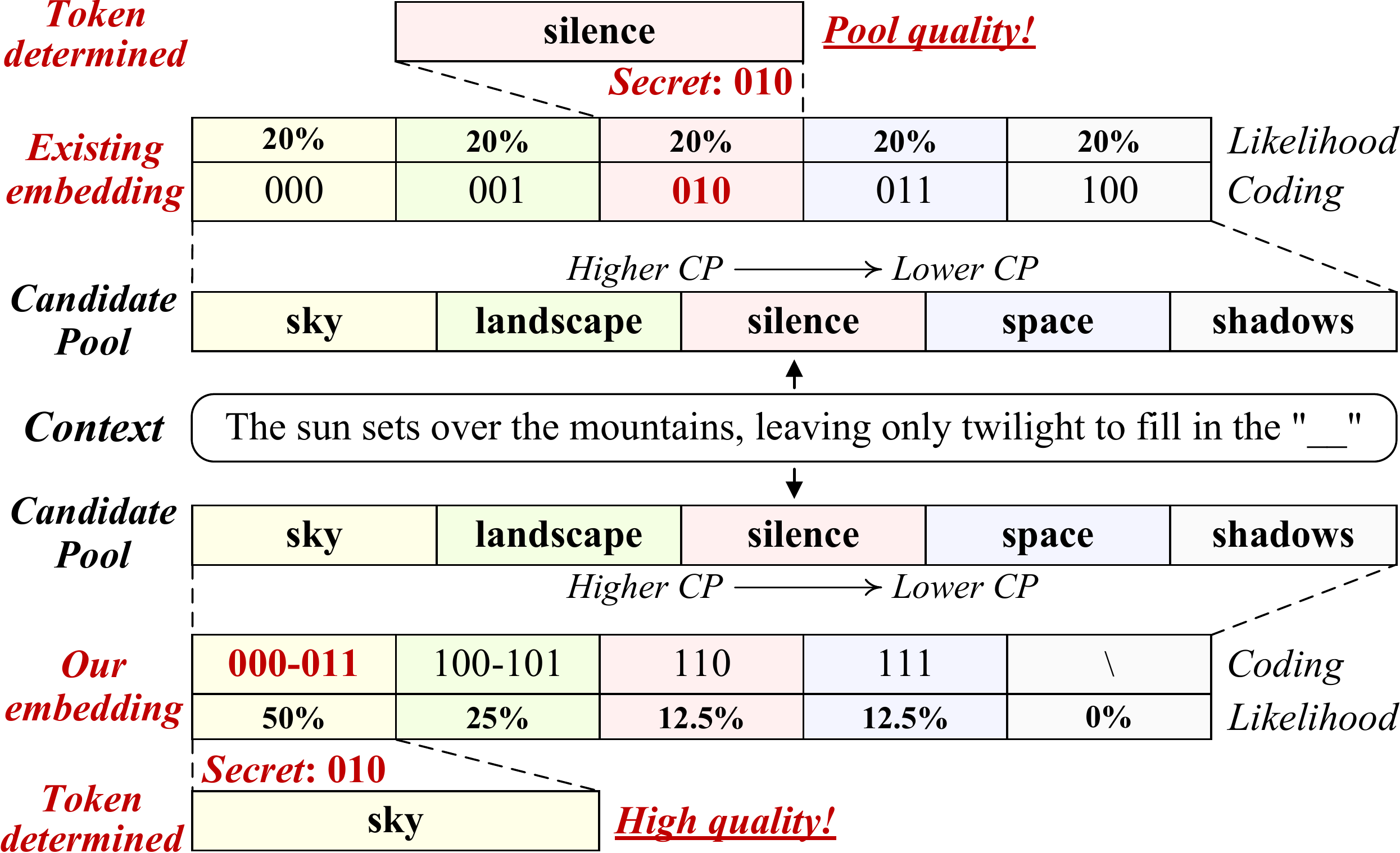}
	\caption{Examples of existing embedding ways and the proposed embedding way. After selecting $n$ tokens with larger CPs to build a candidate pool, the existing way assigns a coding to each token in the candidate pool. Since the secret binary can be arbitrary, to embed the secret, the existing embedding way selects all tokens with the same likelihood of $1/n$. The proposed way will build a roulette area and allocate it according to the CP value to ensure that the larger the CP of the token, the larger the allocation area. Therefore, the proposed way has a higher likelihood of selecting a better token. In the end, the quality of the token determined by the proposed way is generally better than that of the existing ways.}
	\label{example}
\end{figure}

However, the embedding ways of almost all existing schemes allocate one or the same number of codings to the tokens in the candidate pool. That is, when embedding, the likelihood of selecting all tokens is the same. They do not consider that the CP of a portion of the tokens is not high. If the secret is determined to these tokens, the quality of stego generation will be dropped. The example is shown in the Figure \ref{example}.

\begin{figure}[!htbp]
	\centering
	\includegraphics[width=0.41\textwidth]{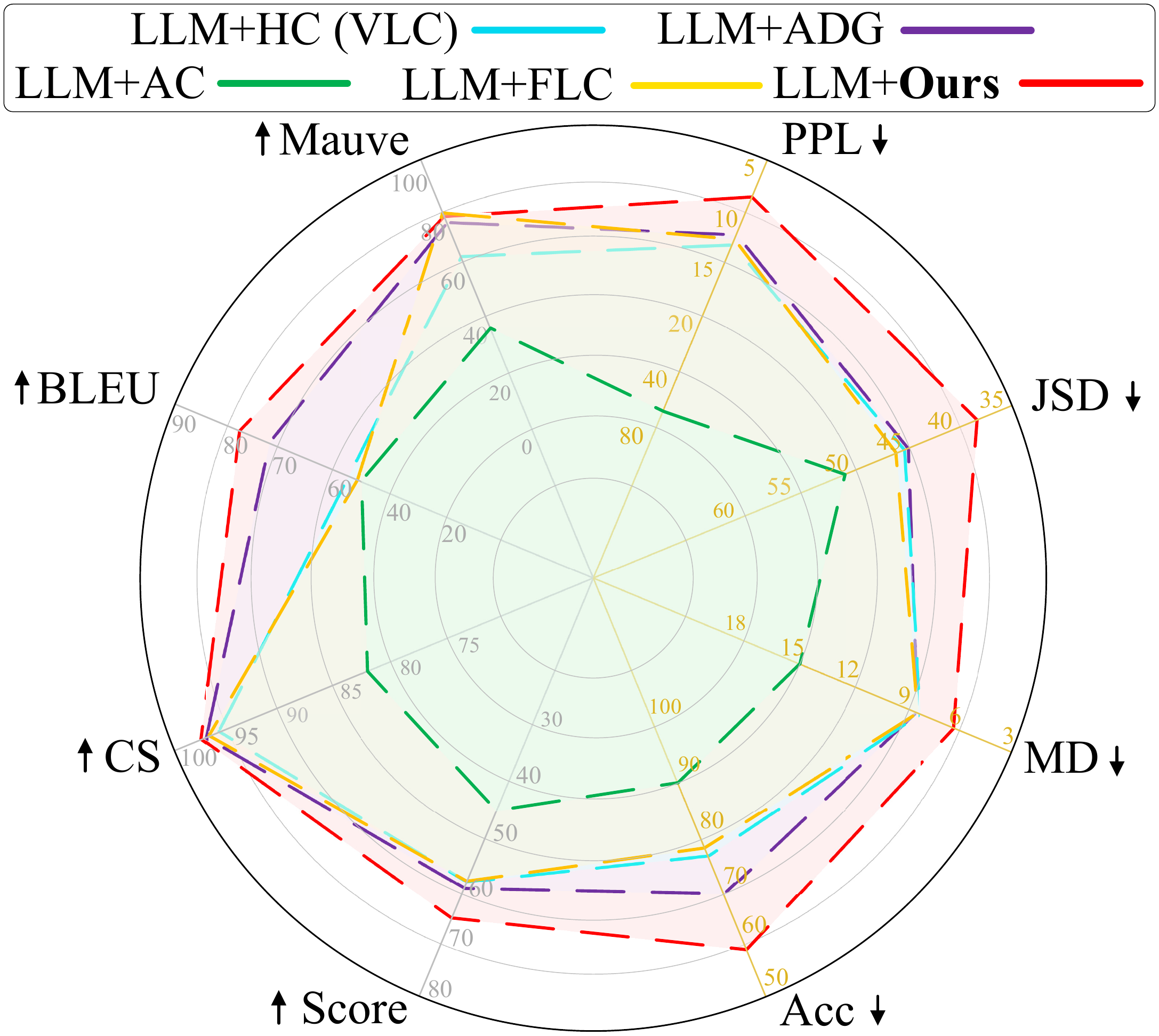}
	\caption{The result of large language model (LLM) + embedding ways. ``$\uparrow$'' and ``$\downarrow$'' represent the higher / lower the value, the better the result. The metrics are found in ``Section \ref{sec31}''. The values are seen in Table \ref{embedding}.}
	\label{intro}
\end{figure}

\noindent Tokens with low CPs will reduce the quality of stego and cause significant differences in its statistical characteristics from cover. Since steganalysis detects stegos by capturing these differences \cite{FEFT, tmode}, poor-quality stego is easier to identify and has weaker anti-steganalysis capabilities. The performance of stegos obtained by different embedding ways is shown in Figure \ref{intro}. Therefore, it is necessary to explore a difficult-to-perceive and detect scheme with concealed embedding. 

This paper proposes the dynamically allocated interval-based steganography, called DAIRstega. Different from the existing schemes, DAIRstega introduces the idea of the roulette wheel to design the embedding way. We believe that the quality of the tokens in the candidate pool is different. Thus, the tokens need to be given different shoot likelihoods by secret, increasing the likelihood of selecting tokens with higher CP. Then, we explore the characteristics of the CP of tokens, and design a variety of relationships between CPs and the number of allocated codings. By analyzing the adaptability of these allocated functions to the invisible information-hiding tasks, the allocation expression is condensed. In addition, DAIRstega can receive instructions and directly control the stego generation.

DAIRstega is built upon the MindSpore framework and the contributions are summarized as follows.

\begin{itemize}	
	\item To augment the concealment of stegos, DAIRstega introduces the idea of the roulette wheel, and dynamically allocates the number of codings. The functions between CPs and allocated numbers are also given. According to Occam's razor, the functions are condensed.
	
	\item To ensure the adaptability of the functions designed, we gave three constraints. 
	
	\item To make the evaluation results consistent with the LLMs era, the high-quality corpus was built and used this corpus as a cover to train the model of all schemes.
	
	\item To verify the effectiveness of the proposed embedding way and DAIRstega, we performed experiments regarding perceptual, statistical, and semantic concealment, as well as anti-steganalysis performance.
	
	\item Experiment verified that DAIRstega can also be regarded as a secure watermarking scheme with high quality, consistency, and unforgeability.
\end{itemize}


\section{Related work}\label{Appendix_A}

The concealment of stego determines the success of covert communication to a great extent. The concealment is primarily manifested in three aspects: perceptual, statistical, and semantic. 

``\underline{Perceptual concealment}" focuses on ensuring that the scheme generates stego with complete and fluent. This is the most basic requirement of steganography. In pursuit of this objective, Fang et al. \cite{TinaFang} proposed an LSTM-based steganography scheme. This scheme segments the vocabulary into several sets based on bit blocks. It then selects tokens with the highest probability that corresponds to the secret information from the candidate pool. Ding et al. \cite{Ding2023TCDS} combined the conditional generation strategy with the replacement technique, using text sequences as auxiliary data in the stego-generation process to enhance the embedding capabilities. Yang et al. \cite{RNN-Stega} trained a language model to generate stegos. Using HC-based (Huffman coding) and FLC-based (Fixed length coding) to encode the candidate pool, these stegos have excellent completeness and fluentness. 

``\underline{Statistical concealment}" requires the stegos distribution to be close to the covers. To achieve this goal, Yang et al. \cite{VAE-Stega} designed an encoder-decoder structure, and VLC-based (Variable length coding) is used to encode the CP. To further reduce the distribution difference between cover and stego, Zhang et al. \cite{ADG} used adaptive dynamic grouping coding to embed the secret recursively. Zhou et al. \cite{GAN-APD} used a generative adversarial network (GAN) to design an adaptive probability distribution steganography. These ensure statistical concealment. 

``\underline{Semantic concealment}" aims to generate a stego that is coherent and semantically controllable. To this end, Li et al. \cite{Topic} put forward a steganography scheme based on the knowledge graph. Yang et al. \cite{Yang2023TDSC} utilized semantic information encoding to embed secret information, realizing the effect of maintaining semantics and increasing the embedding capacity during the translation process. Wang et al. \cite{Hi-Stega} leveraged the relevance of social network context to enhance contextual semantic relevance while maintaining embedding rates. Lu et al. \cite{Lu2023IJCNN} used contextual learning and GPT2 guidance to generate stegos.

\section{Methodology}\label{sec2}

\subsection{Modeling}\label{sec21}
\begin{figure}[!htbp]
	\centering
	\includegraphics[width=0.48\textwidth]{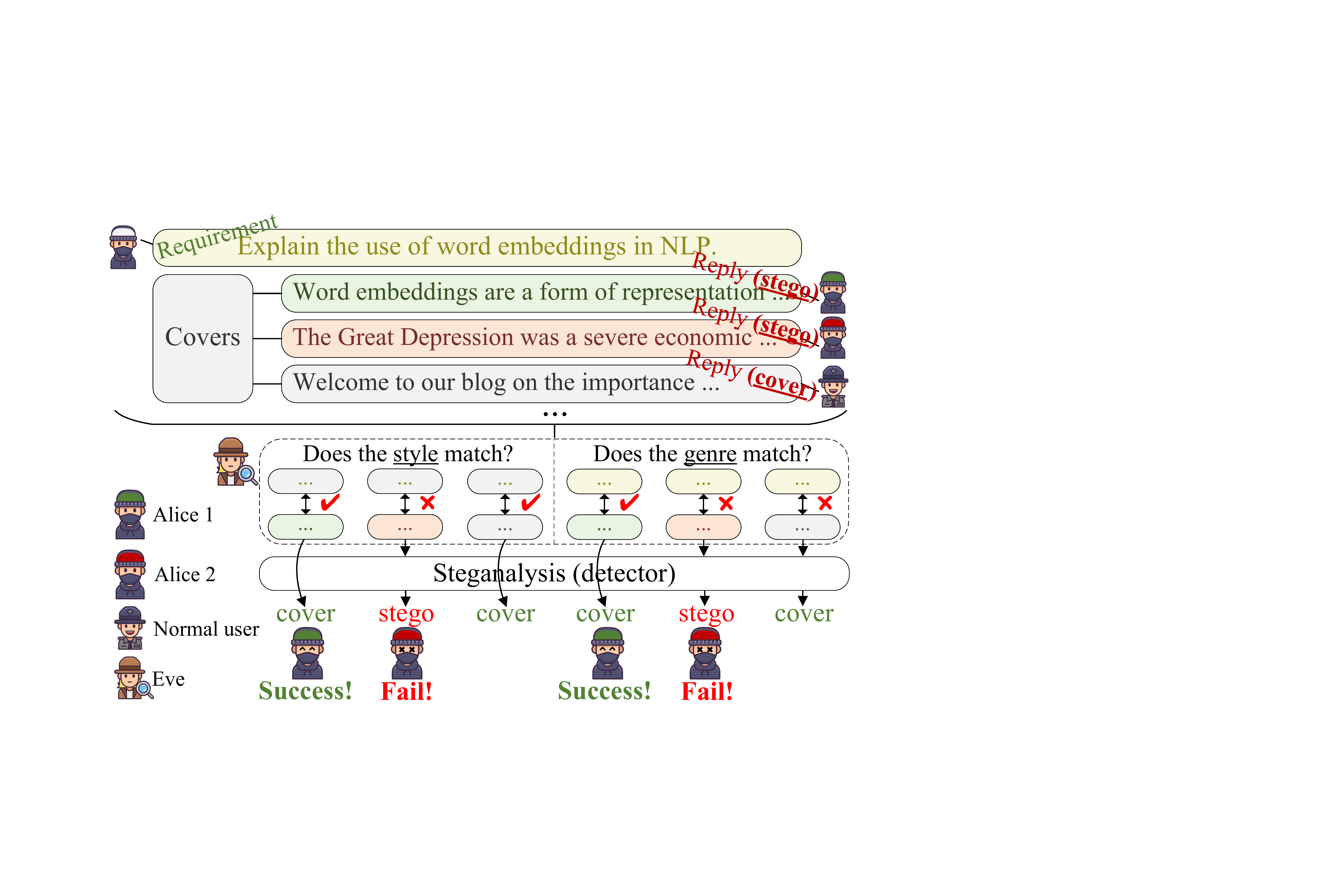}
	\caption{Example of texts being transmitted and detected. Responses that match the discourse are difficult to perceive, thus reducing the risk of steganalysis, such as Alice 1.}
	\label{demo}
\end{figure}

For more covert communications, Alice needs to ensure that the discourse (the style of humans and the genre of texts) is the same as the covers, as shown in Figure \ref{demo}.

The covert communication hopes to avoid Eve's perception and detection and needs to satisfy:

\begin{equation}\label{eq_modeling}
\left\{ \begin{array}{l}
d({P_C},{P_S}) \le \varepsilon \\
C \Rightarrow \mathcal{C}_{style} \sim {P_{\mathcal{C}1}},{S_k} \Rightarrow \mathcal{S}_{style} \sim {P_{\mathcal{S}1}}\\
C \Rightarrow \mathcal{C}_{genre} \sim {P_{\mathcal{C}2}},{\rm{ }}{S_k} \Rightarrow \mathcal{S}_{genre}\sim{P_{\mathcal{S}2}}\\
(d({P_{\mathcal{C}1}},{P_{\mathcal{S}1}}) \le {\varepsilon _1}) \wedge (d({P_{\mathcal{C}2}},{P_{\mathcal{S}2}}) \le {\varepsilon _2})
\end{array} \right.,
\end{equation}

\noindent where, ${P_C}$ and ${P_S}$ are the distribution of cover and stego, $d(\cdot)$ is the difference, $\varepsilon$ is a very small value greater than 0. $\sim$ is following a certain distribution. The above assumptions cover most situations where the secret is transmitted. Complying with the Formula \ref{eq_modeling} can improve the concealment and ensure the conduct of covert communication. The results of this aspect are shown in Table \ref{com_dm}.


\subsection{DAIRstega Overall}\label{sec22}

\begin{figure*}[!htbp]
	\centering
	\includegraphics[width=0.99\textwidth]{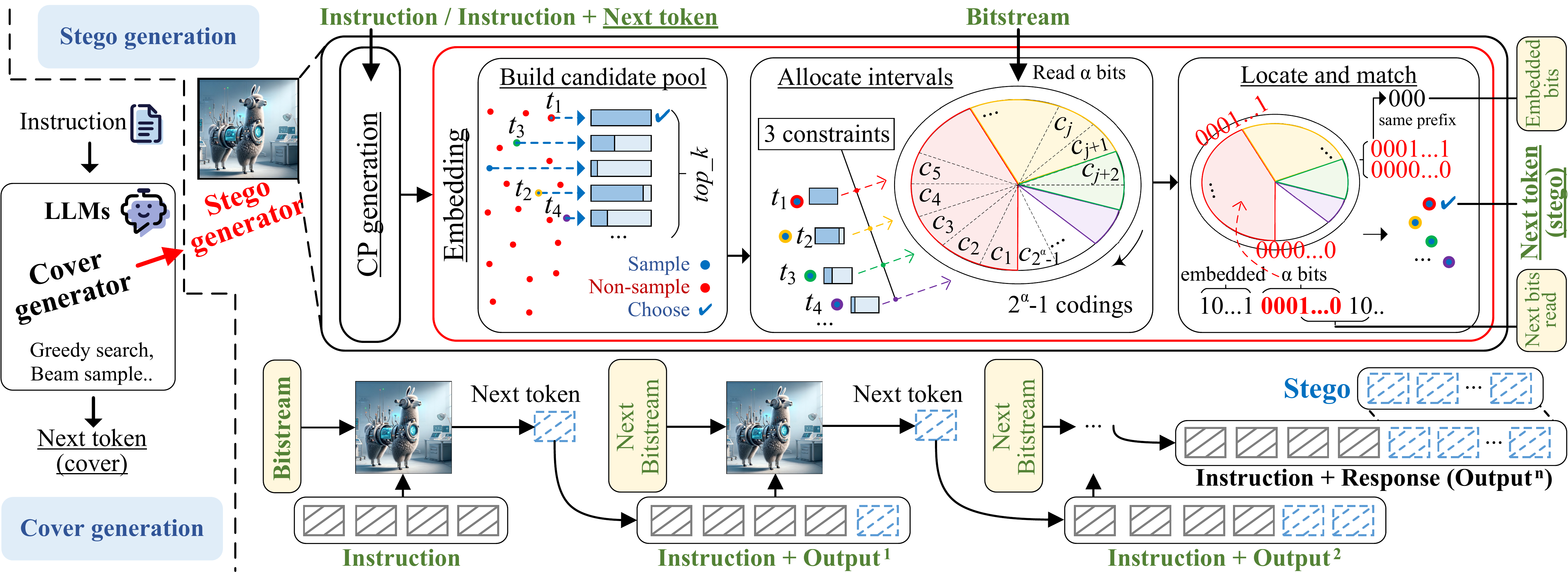}
	\caption{The DAIRstega's framework. The ``stego generation process'' consists of two parts: the ``CP (conditional probability) generation'' and ``Embedding'' modules. Different from existing works, DAIRstega uses the idea of the non-uniform roulette wheel to dynamically allocate different roulette areas (numbers of codings) to tokens in the candidate pool. In this way, the secret is more likely to fall on the tokens with larger CPs, ensuring the quality of stegos. In addition, DAIRstega can not only receive secret information (in the form of bitstreams), but also input the instruction, and finally generate stego that conforms to the instruction.}
	\label{frame}
\end{figure*}

This scheme can generate stegos that are strongly concealed and controllable. It can satisfy the extension of covert communication, i.e., Formula \ref{eq_modeling}. To make the generated stegos more consistent with the CP distribution of covers, DAIRstega uses the idea of the roulette wheel to allocate tokens dynamically in the candidate pool to coding intervals of different lengths. That is, the larger the CP is, the larger the roulette area will be allocated, and the greater the shoot that this token will be selected by secret. Instructions can be public topics, posted questions, or other requirements. Figure \ref{frame} shows the DAIRstega's framework. 

\subsection{Details}\label{sec23}
\paragraph {\textbf{Stego generation process}} It is composed of the ``CP (conditional probability) generation'' ($CPG$) and ``Embedding'' ($Em$) modules. The process formulas are as follows:

\begin{equation}\label{eq_em}
\begin{aligned}
\left\{
\begin{array}{l}
CPG: LM(I) \Rightarrow \mathbf{P} \\
Em: \text{locate}\; \{{m_t}\; \text{in}\; DAI(BC(\mathbf{P}))\}\Rightarrow i,\\
\quad\quad\;\ \text{match}\;  \{i(R_{begin})\; \text{and}\; i(R_{end})\}\\
i = Em(CPG(I))
\end{array}
\right.,
\end{aligned}
\end{equation}

\noindent where, $I$ and $\mathbf{P}$ are the index sequence of a given text and the CP distribution of the next token, $\mathbf{P} = \left[{p_1}, \cdots, {p_v}\right]$, ${p_v}$ is the CP of the $v$-th token in the vocab. ``$LM( \cdot )$'' is the CP calculation process. ``$DAI( \cdot )$'' is the dynamically allocated interval process. ``$BC( \cdot )$'' is the build candidate pool process. ``$\text{locate}\; \{a \;\text{in}\; \circ\}\Rightarrow i $'' means that $i$ is the next token determined by $a$. ``$\text{match}\;  \{i(R_{begin})\; \text{and}\; i(R_{end})\}$'' is the same prefix that matches the beginning and end of $i$'s range.

\paragraph {\textbf{CP generation module}} The input is the instruction or the instruction with part of the subsequent token generated, and the output is the CP distribution $\mathbf{P}$ of the next token.

\paragraph {\textbf{Embedding module}} To ensure that the embedding does not excessively affect the stego quality. In our work, low-CP tokens should not be added to the candidate pool. We sort and intercept the first \textit{top\_k} tokens with higher CPs to build the candidate pool. The $\mathbf{P}$ normalization $\mathbf{P}'=[{p'_1}, \cdots, {p'_{top\_k}}]$ of tokens in candidate pool is:

\begin{equation}\label{eq_candidate pool}
p_j'=\frac{p_j}{\sum\nolimits_j^{top{\rm{\_}}k} {p_j}}, p'_j\in\mathbf{P}', p_j\in\mathbf{P}.
\end{equation}

Then, the $\alpha$-bit secret bitstream is read to determine the total number ($2^\alpha-1$) of intervals. The different numbers of consecutive codings are allocated to tokens in the candidate pool. Intuitively, the allocated relationship between $\mathbf{P}'$ and the number of allocated codings $\mathbf{N}=\left[{n_1}, \cdots, {n_{top\_k}}\right]$ can be linear, the linear function is:

\begin{equation}\label{eq_linear}
n_j= \lfloor \beta p'_j \times ({2^\alpha } - 1)\rceil, n_j\in\mathbf{N}, 
\end{equation}

\noindent where, $\beta$ is the coefficient, $\left\lfloor\cdot\right\rceil$ is the rounding operation. After encoding, the number of $n_j$ is proportional to $p'_j$. It also ensures that the likelihood of selecting a larger $p'_j$ token is greater, directly ensuring the stego quality. To accurately recover the secret, the bits embedded are not the $\alpha$ bits at this moment, but the same prefix at the beginning and end of this interval. The next $\alpha$-bit secret read starts with the next bit embedded this time. 

Further exploration and discovery, if the linear function is used, the payload (unit: bit per word) can be improved. Tokens with very high $p$ will be allocated to most codings, and a large interval length will result in a short prefix of this interval regardless of how many bits are read at a time. Longer stego needs to be generated to embed the given bits, resulting in a lower payload. So, we try to design the nonlinear functions to improve the payload. Using logarithmic and exponential operations, a variety of nonlinear allocations can be designed. However, not all nonlinear allocations are suitable for information hiding. We define three constraints that the functions need to meet. $x$ is a variable that meets all CPs. Since the CP value of the token in the candidate pool is relatively large, $\epsilon^+$ is a smaller value greater than 0, such as 0.1, 0.2, and so on. Although the CP distribution is discrete, the independent variable of the distribution function is continuous, so it is expressed here using derivatives.

\noindent\textbf{Constraint 1:} \textit{(Guaranteed text diversity)}
\begin{equation}\label{eq_c3}
DAI(x) \ge x, \forall x\in \left[\epsilon^+,1\right).
\end{equation}

\noindent If the steganographic embedding can only select a token with extremely high CP, the generated diversity will be lost, and the existence of the stegos will be easily exposed. This constraint indicates that the number of codings allocated to tokens with lower CPs should be appropriately increased to enhance their selection likelihood.

\noindent\textbf{Constraint 2:} \textit{(Guaranteed text quality)}

\begin{equation}\label{eq_c1}
{\frac{{\partial\ {DAI}{(x)}}}{{\partial\ x}}} \ge \epsilon^+, \forall x\in \left[\epsilon^+,1\right),
\end{equation}

\noindent The existing embedding ways assign the same number of codings to all tokens, which results in the same likelihood of randomly selecting each token, thus affecting the generation quality. Therefore, this constraint requires that tokens with larger CP need to be assigned a longer interval, that is, the number of allocations needs to increase with the increase of CP, and the change rate of the number of allocations cannot be too small. If ${\textstyle{{\partial \; DAI(x)} \over {\partial \;x}}} \le \epsilon^+$, which means the number of allocation tokens is similar. It will degenerate into a similar allocation of the same number of codings as the existing work, thereby increasing the likelihood of selecting poor tokens.

\noindent\textbf{Constraint 3:} \textit{(Guaranteed to facilitate embedding)}

\begin{equation}\label{eq_c2}
{\frac{{\partial\ {DAI}{(x)}}}{{\partial\ x}}} \le {\frac{{\partial\ x}}{{\partial\ x}}} \cap {\frac{{\partial^2 {DAI}{(x)}}}{{\partial x^2}}} \le 0, \forall x\in \left[\epsilon^+,1\right).
\end{equation}

\noindent Steganography requires an algorithm that is conducive to embedding information. If you just want high-quality tokens to be selected and blindly increase the interval length of tokens with high CPs, the same prefix at the beginning and end of the interval will be shortened or even non-existent. This is not conducive to embedding secrets and deviates from the basic function of steganography. Therefore, in order to ensure the payload, it is necessary to avoid high CP tokens occupying too much interval length, and the growth trend of ${DAI}{(x)}$ should be gradually slow, that is, mathematically, it can be achieved by limiting the second-order derivative to be less than 0. When the conditions are equal, the corresponding function will be transformed into a linear function.

Here, we give several nonlinear functions:

\begin{equation}\label{eq_nonlinear}
{n_j} = \left\{ {\begin{array}{*{2}{c}}
	{\lfloor {\sqrt[2]{{p_j}'} \times ({2^\alpha } - 1)} \rceil }\\
	{\lfloor {(1 - {e^{ - 2{p_j}'}}) \times ({2^\alpha } - 1)} \rceil}\\
	{\lfloor {{\textstyle{{{{\log }_2}\left( {p_j}' \right) + b} \over b}} \times ({2^\alpha } - 1)} \rceil,b \ge 2}
	\end{array}{\rm{ }}} \right..
\end{equation}

According to the principle of Occam's razor, if you can simplify, you can do less complicated things. We condense the linear and nonlinear functions into a ``simple expression'' of dynamically allocated intervals that satisfy the above three constraints, as shown below:

\begin{equation}\label{eq4}
n_j = \lfloor(\frac{p'}{\sum{p'}})^\beta \times ({2^\alpha } - 1)\rceil, \beta \in (0,1), \alpha \in \mathbb{Z}^+.
\end{equation}

\noindent The smaller $\beta$ is, the smaller the interval of tokens with larger CP will be, while still satisfying the premise that the interval of tokens with larger CP is larger. It will increase the length of the prefix, while reducing the length of stegos to be generated, improving the payload.

\textbf{Please note} that these functions are not fixed, here are just a few examples that meet the constraints. These constraints are not exhaustive, and the final condensed expression is also for simplicity and easy to understand. This can just provide thinking for subsequent research that is more conducive to hiding information.

Loop the stego generation, and stego will be obtained. Algorithm \ref{Alga} illustrates the embedding process.

\begin{algorithm}[!h]
	\caption{Secret embedding of DAIRstega.} 
	\label{Alga}
	\begin{algorithmic}[1]
		\REQUIRE $B = \{x_0,x_1, \cdots ,x_m\} ,x \in \{ 0,1\}$; Instruction.
		\ENSURE Stego $S = [{t_{n + 1}},{t_{n + 2}},\cdots ,{\rm{<EOS>}}]$.
		\WHILE{Not the end of the bitstream}
		\STATE Map Instruction $\Rightarrow{I_1} = [{i_1},{i_2}, \cdots ,{i_n}]$;
		\WHILE{Not $I_{<\text{EOS}>}$ in $I$}
		\STATE Get $\mathbf{P}$ of $i_k, k \in [n+1,n+2,\cdots,{\rm{<EOS>}}]$;
		\STATE Build candidate pool and get $\mathbf{P'}$ by Formula \ref{eq_candidate pool};
		\STATE Read the unembedded $\alpha$-bit bitstream;
		\STATE Give $\beta\Rightarrow n_j$ by Formula \ref{eq4};
		\STATE Allocate $\mathbf{N}$ to $I$;
		\STATE Determine the interval by the $\alpha$ bits;
		\ENDWHILE
		\STATE $\text{Locate}\; \{{\alpha\text{-bit} \ B}\; \text{in}\; \text{interval}\}\Rightarrow i_k$;
		\STATE $\text{Match}\;  \{i_k(R_{begin})\; \text{and}\; i_k(R_{end})\} \Rightarrow$ embedded secret at this moment;
		\ENDWHILE
		\STATE Map $[{i_{n + 1}}, \cdots ,{i_{{\rm{ < EOS > }}}}]\Rightarrow [{t_{n + 1}}, \cdots ,{\rm{<EOS>}}]$.
	\end{algorithmic} 
\end{algorithm}

\paragraph {\textbf{Secret extraction process}} Bitstream extraction and embedding are a pair of inverse operations. Same as Discop \cite{Discop}, we also need to use the same scheme, initial context (instruction), and stegos. The instructions remain requirements visible to everyone in social networks. The formula of the extraction process $Ex$ is as follows: 

\begin{equation}
\left\{ \begin{array}{l}
Ex:\text{locate}\;\ \{{i_{n + 1}} \;\ {\rm{in}} \;\ DAI(BC(P))\},\\
\quad\quad\; \text{match}\;  \{i(R_{begin})\; \text{and}\; i(R_{end})\}\\
B_j = Ex(CPG(I)), B_j \in B
\end{array} \right..
\end{equation}

\noindent The interval is located by the stego's next token at this moment, and the same prefix is the currently extracted secret.


\section{Experiments}\label{sec3}
To ensure fairness and reliability in comparisons between schemes, each result was repeated 5 times and averaged to provide the results. All experiments are run on the Linux Server with 24G NVIDIA GeForce RTX 4090 GPUs. 


\subsection{Settings}\label{sec31}
\paragraph {\textbf{Dataset}} We obtained a large amount of data from Wikipedia, Twitter, GPT4 \cite{GPT4}, and the publicly available data for fine-tuning LLMs\footnote[1]{ https://github.com/tloen/alpaca-lora/tree/main}. The final dataset contains 57,414 texts and nearly 100 different discourse characteristics to fine-tune the language models of all schemes.

\paragraph {\textbf{MindSpore framework}} All experiments are based on the MindSpore framework. MindSpore is an open-source AI framework introduced by Huawei, designed to support flexible deployment across cloud, edge, and device scenarios. In this research, MindSpore played a significant role in unleashing the computational power of the hardware.

\paragraph {\textbf{Model configuration}} LLM is divided into closed-source and open-source LLMs, closed source such as Gemini, GPT4, etc., open-source such as LLaMA2-7B, etc. For detailed information on closed-source LLMs, please refer to the review of comprehensive research in related fields \cite{G1, G3}. Since DAIRstega needs to modify the token sampling inside LLMs, this paper uses open-source LLMs for experiments. DAIRstega can access any open-source LLMs.  In the comparison experiments, the LLaMA2-7B model is used \cite{LLaMA2} as the base LLM. The LLaMA2-7B and LLaMA2-13B models are used in the ablation experiments of the different LLMs, and all models are operated with 16-bit precision. For simplicity, we use $\beta=1,0.5$ and $\alpha=8,32,48$ to perform DAIRstega.

\paragraph {\textbf{Baselines}} Since the steganography scheme consists of a language model and an embedding way, the language models of existing steganography are different. Therefore, to ensure fairness in the comparison, we not only compared the schemes, but also compared the performance of different embedding ways. That is to say, comparison is divided into two parts: one is the comparison between the proposed embedding way and the existing embedding ways, and the other is the comparison between DAIRstega and the prevalent works in recent years.

\begin{table*}[!htbp]
	\centering
	\footnotesize
	\setlength{\tabcolsep}{0.25mm}
	\caption{Overview of each part of the experiments. It includes: experimental part (Focus), Section, Metrics, Base LLM, Table, and supplementary materials for table (Supplement).}\label{Overview}
	\begin{tabular}{cc||ccccc}
		\toprule[1.2pt]
		\multicolumn{2}{c||}{Focus} & Section & Metrics&Base LLM& Table & Supplement \\
		\midrule
		\multirow{6}{*}{Comparison study} & Embedding ways & \ref{sec32} & Perceptual, Statistical, Semantic, Anti-steganalysis &\multirow{6}{*}{LLaMA2-7B}&\ref{embedding}& \ref{app_embedding1}, \ref{app_embedding2}\\
		& \multirow{4}{*}{Steganography baselines}  &\ref{sec331}& Perceptual&& \ref{com_tq} & -- \\
		&  &\ref{sec322}& Statistical&& \ref{com_sa} & -- \\
		&  &\ref{sec323}& Semantic&& \ref{com_dm} & \ref{app_t4} \\
		&  &\ref{sec324}& Anti-steganalysis&& \ref{com_as} & \ref{app_t5}, \ref{llm-stega} \\
		&Related-task baseline  & \ref{sec34} & Quality, Consistency, Unforgeability&&\ref{com_watermark} & -- \\
		\midrule
		\multirow{2}{*}{DAIRstega exploration} & Longer stegos & \ref{sec35} & Perceptual, Semantic&LLaMA2-7B&\ref{tab5}& \ref{cm} \\
		& Ablation of LLMs & \ref{sec36} & Perceptual, Statistical, Semantic, Anti-steganalysis&LLaMA2-7B, 13B& \ref{ae}& -- \\
		\bottomrule[1.2pt]
	\end{tabular}%
\end{table*}%

\begin{table*}[!b]
	\centering
	\small
	\setlength{\tabcolsep}{2.7mm}
	\caption{The overall results of different embedding ways (based on MindSpore framework). The embedding rate of each way is near 1Bpw (bit per word). \textbf{Bold} and ``\underline{ *}" are the best and the suboptimal result. The ``LLM"s in this table are all LLaMA2-7B. For the steganography works involved in these embedding ways, see ``Section \ref{sec31}". ``$\uparrow$" is that the higher the value, the better the result. ``$\downarrow$" means the lower the value, the better the result. The complete data can be seen in Table \ref{app_embedding1} and Table \ref{app_embedding2} in Appendix \ref{Appendix_B}.}\label{embedding}%
	\begin{tabular}{c||ccccccccc}
		\toprule[1.2pt]
		\multirow{2}{*}{\textbf{Embedding ways}}&\multicolumn{2}{c|}{Perceptual}&\multicolumn{5}{c|}{Statistical}&\multicolumn{2}{c}{Semantic $\hookleftarrow$}\\  
		& PPL $\downarrow$ & \multicolumn{1}{c|}{$\Delta$Pcs $\downarrow$}&CS $\uparrow$  & JSD $\downarrow$ & ED $\downarrow$&MD $\downarrow$& \multicolumn{1}{c|}{$\Delta$DP $\downarrow$}&Mauve $\uparrow$ &BLEU $\uparrow$\\
		\midrule[0.5pt]
		LLM+HC& 13.331  & \multicolumn{1}{c|}{5.639}  & 96.54  & 45.79  & 0.263  & \underline{8.255*}  &  \multicolumn{1}{c|}{3.46}  & 70.60 & 54.26 \\
		LLM+ADG &  \underline{12.750*}  & \multicolumn{1}{c|}{\underline{5.058*}}
		& \underline{97.20*}  & \underline{45.59*}  & \underline{0.240*}  & 8.316   & \multicolumn{1}{c|}{\underline{2.80*}}  & 77.86 & \underline{71.79*} \\
		LLM+AC & 73.454 & \multicolumn{1}{c|}{65.762}  & 82.16 & 51.64 & 0.746 & 15.415 &  \multicolumn{1}{c|}{17.84} & 35.64 & 50.92\\
		LLM+FLC & 12.883  & \multicolumn{1}{c|}{5.190}  & 97.04  & 46.56  & 0.241  & 8.280  &  \multicolumn{1}{c|}{2.96}  & \textbf{79.48}& 54.88 \\
		LLM+\textbf{Ours} & \textbf{8.030} &	\multicolumn{1}{c|}{\textbf{0.342}}  &	\textbf{97.43} &	\textbf{38.60} &	\textbf{0.227} &	\textbf{6.341}  &\multicolumn{1}{c|}{\textbf{2.57}} &\underline{78.53*}&	\textbf{77.42}	\\
		\midrule[1.2pt]
		\multirow{3}{*}{\textbf{Embedding ways}}&\multicolumn{1}{c|}{$\hookrightarrow$}&\multicolumn{8}{c}{Anti-steganalysis}\\
		& \multicolumn{1}{c|}{\multirow{2}{*}{Score $\uparrow$}}& \multicolumn{2}{c}{LS\_CNN \cite{LS_CNN}} &\multicolumn{2}{c}{TS\_CSW \cite{TS_CSW}}&\multicolumn{2}{c}{EILG \cite{EILG}}&\multicolumn{2}{c}{UP4LS \cite{UP4LS}}\\
		&\multicolumn{1}{c|}{}&Acc $\downarrow$ & F1 $\downarrow$&Acc $\downarrow$& F1 $\downarrow$&Acc $\downarrow$& F1 $\downarrow$&Acc $\downarrow$& F1 $\downarrow$\\
		\midrule[0.5pt]
		
		LLM+HC&	\multicolumn{1}{c|}{58.44} &	77.62 & 78.41 & 78.09 & 78.25 & 78.58 & 78.29 & 86.73 & 86.13\\
		LLM+ADG&	\multicolumn{1}{c|}{\underline{59.18*}}&\underline{70.10*} & \underline{70.17*} & \underline{69.45*} & \underline{69.34*} & \underline{70.99*} & \underline{70.49*} & \underline{84.00*} & \underline{84.23*} \\
		LLM+AC &	\multicolumn{1}{c|}{45.10}&90.15 & 89.98 & 89.69 & 89.37 & 86.92 & 86.52 & 93.62 & 93.37 \\
		LLM+FLC  &	\multicolumn{1}{c|}{58.44} &	78.53 & 80.05 & 82.72 & 83.32 & 76.67 & 76.15 & 89.03 & 89.20 \\
		LLM+\textbf{Ours}&	\multicolumn{1}{c|}{\textbf{65.41}} &	\textbf{58.95} &\textbf{58.74}& \textbf{55.28} &	\textbf{61.69} &	\textbf{65.01} &	\textbf{65.83} &	\textbf{72.92} &	\textbf{73.30}  \\
		\bottomrule[1.2pt]
	\end{tabular}%
	\begin{tablenotes}
		\item[] \footnotesize 1. HC (Huffman coding) is used in VAE-Stega \cite{VAE-Stega}, LSCS \cite{Lu2023IJCNN}, RNNstega \cite{RNN-Stega}, GAN-APD \cite{GAN-APD}, JLS \cite{Ding2023TCDS}, and Discop \cite{Discop}.
		\item[] 2. ADG (Adaptive dynamic grouping) is used in Hi-Stega \cite{Hi-Stega} and ADG \cite{ADG}.
		\item[] 3. AC (Arithmetic coding) is used in VAE-Stega \cite{VAE-Stega}, Hi-Stega \cite{Hi-Stega}, and Meteor \cite{Meteor}.
		\item[] 4. FLC (Fixed length coding) is used in RNNstega \cite{RNN-Stega} and JLS \cite{Ding2023TCDS}.
	\end{tablenotes}
\end{table*}%

$\cdot$ In terms of the \underline{embedding ways}, the existing works can be summarized as follows: (1) Huffman coding (HC) (or Variable length coding (VLC)), (2) Adaptive dynamic grouping (ADG), (3) Arithmetic coding (AC), and (4) Fixed length coding (FLC). \textbf{(1) HC} is used in VAE-Stega \cite{VAE-Stega}, LSCS \cite{Lu2023IJCNN}, RNNstega \cite{RNN-Stega}, GAN-APD \cite{GAN-APD}, JLS \cite{Ding2023TCDS}, Discop \cite{Discop}. \textbf{(2) ADG} is used in Hi-Stega \cite{Hi-Stega} and ADG \cite{ADG}. \textbf{(3) AC} is used in VAE-Stega \cite{VAE-Stega}, Hi-Stega \cite{Hi-Stega}, and Meteor \cite{Meteor}. \textbf{(4) FLC} is used in RNNstega \cite{RNN-Stega}, JLS \cite{Ding2023TCDS}.

$\cdot$ In terms of the \underline{prevalent works}, we selected the steganography and watermarking works that showed excellent performance as the baselines. \textbf{Steganography-task baselines:} \textbf{(1) Fang} \cite{TinaFang}. \textbf{(2) RNNstega} \cite{RNN-Stega}. \textbf{(3) ADG} \cite{ADG}. \textbf{(4) LSCS} \cite{Lu2023IJCNN}. \textbf{Related-task baseline:} \textbf{(5) PLMmark} \cite{Liu2023AAAI}. It uses pre-trained language models to provide the first secure black-box watermarking scheme. They can compare the generation effects of DAIRstega from different aspects, and these performances have been widely recognized.

\paragraph {\textbf{Evaluation metrics}} $\downarrow$ means the lower the value, the better, $\uparrow$ means the higher the value, the better. 

$\cdot$ In terms of \underline{perceptual concealment} (text quality), we adopt \textbf{(1) PPL $\downarrow$} (stego perplexity). \cite{Wu2023ICML, Ding2023TCDS}. \textbf{(2) $\Delta$Pcs $\downarrow$} (The difference between PPL of cover and stego \cite{VAE-Stega}). 

$\cdot$ In terms of \underline{statistical concealment} (statistical analysis), we adopt \textbf{(3) CS $\uparrow$}. (Cosine similarity). \textbf{(4) JSD $\downarrow$}, which averages the KL divergence. \textbf{(5) ED $\downarrow$} (Euclidean distance). \textbf{(6) MD $\downarrow$} (Manhattan distance). \textbf{(7) $\Delta$DP $\downarrow$} (Dot product difference \cite{Cai2021ICML}). 

$\cdot$ In terms of \underline{semantic concealment} (discourse matching), we adopt \textbf{(8) $\Delta$LDA $\downarrow$}. \textbf{(9) Mauve $\uparrow$} \cite{MAUVE}. Mauve is a metric that considers fluency and consistency. \textbf{(10) BLEU $\uparrow$} (Bilingual evaluation understudy \cite{BLEU}). \textbf{(11) Score (BERTScore) $\uparrow$} \cite{bertscore}. Score considers the similarity at the segment level. $\cdot$ 

In terms of \underline{anti-steganalysis}, steganalysis is essentially a detection method \cite{G2}. It can detect whether there is secret information in the text \cite{RLS-DTS}. we use the high-performance linguistic steganalysis LS\_CNN \cite{LS_CNN}, TS\_CSW \cite{TS_CSW}, EILG \cite{EILG}, and UP4LS \cite{UP4LS}: \textbf{(12) Acc $\downarrow$} (Accuracy) and \textbf{(13) F1 $\downarrow$} (F1-score).

\paragraph {\textbf{Experimental overview}} We provide an overview of each part of the experiments, as shown in Table \ref{Overview}.

\subsection{Comparison with the different embedding ways}\label{sec32}

To illustrate the effect of the embedding way, we performed this experiment. We add the existing works with the same LLMs as ours. The results are shown in Table \ref{embedding}.

Table \ref{embedding} shows that the proposed embedding way can improve the stego quality (perceptual concealment), better simulate the CP distribution of covers (statistical concealment), and get the optimal anti-steganalysis. Overall performance performs better than the existing ways.

\subsection{Comparison with steganography-task baselines}\label{sec33}
\subsubsection{Perceptual concealment (text quality)}\label{sec331}

Table \ref{com_tq} shows the comparison regarding the perceptual concealment.

\begin{table}[!htbp]
	\centering
	\small
	\setlength{\tabcolsep}{2.3mm}
	\caption{Perceptual concealment comparison of stegos generated by DAIRstega and steganography baselines. \textbf{Bold} is the best result. ``\underline{ *}" is the suboptimal result. $\alpha$ and $\beta$ can be found in Formula \ref{eq4}. The meanings of the parameters ``bin", ``bit", and ``$\tau$" are the same in Fang, RNNstega, and ADG schemes.}\label{com_tq}
	\begin{tabular}{ccc||cc}
		\toprule[1.2pt]
		Schemes&\multicolumn{2}{c||}{Param / Bpw}&PPL $\downarrow$&$\Delta$Pcs $\downarrow$\\
		\midrule[0.5pt]
		\multicolumn{1}{c}{\multirow{2}{*}{Fang \cite{TinaFang}}} & \multicolumn{2}{c||}{bin = 1 / 1.00}  & 235.271 & 227.578 \\
		& \multicolumn{2}{c||}{bin = 3 / 3.00} & 269.303 & 261.611 \\
		\midrule[0.5pt]
		
		\multicolumn{1}{c}{\multirow{2}{*}{RNNstega \cite{RNN-Stega}}} & \multicolumn{2}{c||}{bit = 1 / 1.00}& 79.987 & 72.294 \\
		&  \multicolumn{2}{c||}{bit = 3 / 2.63} & 99.302 & 91.610 \\  
		\midrule[0.5pt]     
		
		\multirow{1}{*}{ADG \cite{ADG}}& \multicolumn{2}{c||}{$\tau=0.5$ / 4.38} & 197.498 & 189.806 \\
		\midrule[0.5pt]
		
		LSCS \cite{Lu2023IJCNN}& \multicolumn{2}{c||}{- / 1.12}& 9.248 & 1.556 \\
		\midrule[0.5pt]
		
		\multirow{6}{*}{\textbf{DAIRstega}} & \multirow{3}{*}{$\beta$ = 1} & $\alpha$ = 8 / 1.10 & \textbf{7.686} & \textbf{0.006} \\
		&&$\alpha$ = 32 / 1.11 & \underline{8.184*} & \underline{0.492*} \\
		&&$\alpha$ = 48 / 1.13 & 8.219 & 0.527 \\
		\cdashline{2-5}[3pt/2pt]
		& \multirow{3}{*}{$\beta$ = 0.5} & $\alpha$ = 8 / 1.89 & 13.775  & 6.082 \\
		&&$\alpha$ = 32 / 2.56 & 20.876 & 13.184 \\
		&&$\alpha$ = 48 / 2.58 & 22.069 & 14.377 \\
		\bottomrule[1.2pt]
	\end{tabular}%
\end{table}%

According to the results in Table \ref{com_tq}, it can be found that the stego quality of DAIRstega markedly surpasses that of the baselines. This shows that DAIRstega can be more consistent with the linguistic characteristics of covers, enhancing the perceptual concealment of the stegos.

\subsubsection{Statistical concealment (statistical analysis)}\label{sec322}
Table \ref{com_sa} shows the comparison between DAIRstega and baselines regarding the statistical analysis between covers and stegos.

\begin{table}[!htbp]
	\centering
	\scriptsize
	\setlength{\tabcolsep}{1.8mm}
	\caption{Statistical concealment comparison of stegos generated by DAIRstega and steganography baselines. The meanings of ``\textbf{Bold}", ``\underline{ *}", ``bin", ``bit", and ``$\tau$" are the same in Table \ref{com_tq}.}\label{com_sa}
	\begin{tabular}{ccc||ccccc}
		\toprule[1.2pt]
		Schemes&\multicolumn{2}{c||}{Param}&CS $\uparrow$&JSD $\downarrow$&ED $\downarrow$&MD $\downarrow$&$\Delta$DP $\downarrow$\\
		\midrule[0.5pt]
		
		\multicolumn{1}{c}{\multirow{2}{*}{Fang \cite{TinaFang}}} & \multicolumn{2}{c||}{bin = 1} & 9.31  & 63.43  & 1.347  & 17.911  & 90.69  \\
		&  \multicolumn{2}{c||}{bin = 3} & 9.30  & 63.34  & 1.347  & 17.876  & 90.70  \\            
		\midrule[0.5pt]
		
		\multicolumn{1}{c}{\multirow{2}{*}{RNNstega \cite{RNN-Stega}}} & \multicolumn{2}{c||}{bit = 1} & 95.97  & 41.51  & 0.284  & 7.530  & 4.03  \\
		& \multicolumn{2}{c||}{bit = 3} &  96.32 & 39.38  & 0.271  & 7.371  & 3.68  \\        
		\midrule[0.5pt]
		
		ADG \cite{ADG}& \multicolumn{2}{c||}{$\tau=0.5$}& 96.82 & 43.02  & 0.252  & 8.232  & 3.18  \\
		\midrule[0.5pt]
		
		LSCS \cite{Lu2023IJCNN}&\multicolumn{2}{c||}{-}&  94.20 &40.28 & 0.340 & 7.182 & 5.80\\
		\midrule[0.5pt]
		
		\multirow{6}{*}{\textbf{DAIRstega}} &\multirow{3}{*}{$\beta$=1} & $\alpha$=8& 97.20 & 38.86  & 0.237  & 6.458  & 2.80  \\
		&&$\alpha$=32& \textbf{97.65}  & 38.48 & \textbf{0.217} & \underline{6.293*}  & \textbf{2.35} \\
		&&$\alpha$=48&\underline{97.45*} & 38.46  & \underline{0.226*}  & \textbf{6.273} & \underline{2.55*}  \\
		\cdashline{2-8}[3pt/2pt]
		&\multirow{3}{*}{$\beta$=0.5}& $\alpha$=8&97.39&\textbf{37.80}&0.229 &7.204 &2.61\\
		&&$\alpha$=32&92.43&39.69 &0.389 &8.976 &7.57 \\
		&&$\alpha$=48&92.44&\underline{38.31*} &0.389 &8.592 &7.56\\
		\bottomrule[1.2pt]
	\end{tabular}%
\end{table}%

According to the comparison results in Table \ref{com_sa}, it can be found that DAIRstega can better simulate the statistical distribution of cover, enhancing the \underline{statistical concealment}.

\subsubsection{Semantic concealment (discourse matching)}\label{sec323}
Table \ref{com_dm} shows the comparison regarding semantic concealment.

\begin{table}[H]
	\centering
	\scriptsize
	\caption{Semantic concealment comparison of stegos generated by DAIRstega and steganography baselines. The unit is \%. The complete data can be seen as Table \ref{app_t4} in Appendix \ref{Appendix_C}.}\label{com_dm}
	\begin{tabular}{ccc||cccc}
		\toprule[1.2pt]
		Schemes&\multicolumn{2}{c||}{Param}&$\Delta$LDA $\downarrow$&Mauve $\uparrow$&BLEU $\uparrow$&Score $\uparrow$\\
		\midrule[0.5pt]
		
		\multicolumn{1}{c}{\multirow{2}{*}{Fang \cite{TinaFang}}} & \multicolumn{2}{c||}{bin = 1}  & 11.964 & 2.15 & 0.90 &  42.58 \\
		& \multicolumn{2}{c||}{bin = 3} &11.964 & 1.96 & 0.90&  41.65\\       
		\midrule[0.5pt]
		
		\multicolumn{1}{c}{\multirow{2}{*}{RNNstega \cite{RNN-Stega}}} & \multicolumn{2}{c||}{bit = 1} &  0.062 & 12.73 & 1.87&  52.57\\
		& \multicolumn{2}{c||}{bit = 3} & 18.919 & 11.97& 1.83&  52.36\\
		\midrule[0.5pt]
		
		\multirow{1}{*}{ADG \cite{ADG}} & \multicolumn{2}{c||}{$\tau=0.5$}& 15.901 & 18.54 & 2.72&  47.55\\
		\midrule[0.5pt]
		
		LSCS \cite{Lu2023IJCNN}&\multicolumn{2}{c||}{-}&  0.022 & 2.40& 31.95&59.22\\
		\midrule[0.5pt]
		
		\multicolumn{1}{c}{\multirow{6}{*}{\textbf{DAIRstega}}} &\multirow{3}{*}{$\beta$=1}&$\alpha$=8& \underline{0.017*} & \underline{79.91*}  & 76.37 &  \textbf{65.76} \\
		&&$\alpha$=32&0.022 & 76.97  & \underline{77.85*} & 65.22\\
		&&$\alpha$=48 &0.020 & 78.70  & 76.53 &  \underline{65.25*} \\ 
		\cdashline{2-7}[3pt/2pt]
		
		&\multirow{3}{*}{$\beta$=0.5}&$\alpha$=8&0.018&\textbf{80.83}&73.19&62.49\\
		&&$\alpha$=32&0.029&75.58&66.99&60.89\\
		&&$\alpha$=48&\textbf{0.006}&72.40&\textbf{79.64}&60.29\\  
		\bottomrule[1.2pt]
	\end{tabular}%
\end{table}%

\subsubsection{Anti-steganalysis}\label{sec324}
Table \ref{com_as} shows the anti-steganalysis comparison of DAIRstega and baselines. 

\begin{table}[H]
	\centering
	\scriptsize
	\setlength{\tabcolsep}{0.9mm}
	\caption{Anti-steganalysis (Acc $\downarrow$) comparison of stegos generated by DAIRstega and steganography baselines. The unit is \%. The complete data can be seen as Table \ref{app_t5} in Appendix \ref{Appendix_C}. We also give the detection of the other 7 steganalysis works \cite{FEFT, rbilstmc, Zou, SSLS, sesy, llmsteganalysis, tmode} (including non-BERT, BERT, and LLM-based). It can be seen as Table \ref{llm-stega} in Appendix \ref{Appendix_D}.}\label{com_as}
	\begin{tabular}{ccc||cccc}
		\toprule[1.2pt]
		Schemes&\multicolumn{2}{c||}{Param} &LS\_CNN \cite{LS_CNN}&TS\_CSW \cite{TS_CSW}&EILG \cite{EILG}&UP4LS \cite{UP4LS}\\
		\midrule[0.5pt]
		
		\multicolumn{1}{c}{\multirow{2}{*}{Fang \cite{TinaFang}}} & \multicolumn{2}{c||}{bin = 1} & 99.70 & 88.34 & 99.50  &99.75  \\
		&  \multicolumn{2}{c||}{bin = 3} & 99.75 &  89.03 & 99.52& 99.88\\
		\midrule[0.5pt]
		
		\multicolumn{1}{c}{\multirow{2}{*}{RNNstega \cite{RNN-Stega}}} & \multicolumn{2}{c||}{bit = 1} & 85.81 & 81.39&  85.86 &97.42\\
		& \multicolumn{2}{c||}{bit = 3} & 82.93&  80.35& 82.71 & 96.78\\
		\midrule[0.5pt]
		
		ADG \cite{ADG}& \multicolumn{2}{c||}{$\tau=0.5$}& 85.86& 82.23& 84.86&  98.34\\
		\midrule[0.5pt]
		
		LSCS \cite{Lu2023IJCNN}& \multicolumn{2}{c||}{-}&78.86& 81.19&84.15&  96.22\\
		\midrule[0.5pt]
		
		\multirow{6}{*}{\textbf{DAIRstega}}&\multirow{3}{*}{$\beta$=1}&$\alpha$=8&59.50& 56.82& 67.25& 75.40\\
		&&$\alpha$=32&\underline{59.35*}&  \textbf{54.05} &\textbf{62.28} & 71.97\\
		&&$\alpha$=48&\textbf{58.00} & \underline{54.98*} & \underline{65.51*} &   \underline{71.40*}  \\
		\cdashline{2-7}[3pt/2pt]
		&\multirow{3}{*}{$\beta$=0.5}&$\alpha$=8&62.88 &63.13	&65.67	&\textbf{69.04}	\\
		&&$\alpha$=32&72.06	&68.73	&71.71	&76.41\\
		&&$\alpha$=48&70.17	&69.48	&70.64	&79.46\\
		\bottomrule[1.2pt]
	\end{tabular}%
\end{table}%

\subsection{Comparison with related-task baseline}\label{sec34}
The rise of LLMs has spawned many excellent watermarking papers \cite{Liu2023AAAI, watermark1}. They provide solutions for copyright protection in terms of high text quality \cite{watermark1}, consistency \cite{watermark2}, and unforgeability \cite{Liu2023AAAI}. Excitingly, DAIRstega can also be regarded as an invisible watermarking with high quality, consistency, and unforgeability.

The focus of this paper is not watermarking, we only compare it with the PLMmark, as shown in Table \ref{com_watermark}.

\begin{table*}[!htbp]
	\centering
	\small
	\caption{Quality, consistency, and unforgeability comparison with watermarking baseline. The complete data can be seen in Table \ref{app_t4} and Table \ref{app_t5} in Appendix \ref{Appendix_C}.}\label{com_watermark}
	\begin{tabular}{cc||cc|ccc|ccc}
		\toprule[1.2pt]
		\multicolumn{2}{c||}{\multirow{2}{*}{Scheme}}&\multicolumn{2}{c|}{Quality}&\multicolumn{3}{c|}{Consistency}&\multicolumn{3}{c}{Unforgeability}\\ 
		&&PPL $\downarrow$& $\Delta$Pcs $\downarrow$&$\Delta$LDA $\downarrow$&Mauve $\uparrow$&Score $\uparrow$&TS\_CSW $\downarrow$&EILG $\downarrow$&UP4LS $\downarrow$\\
		
		\midrule[0.5pt]
		\multicolumn{2}{c||}{PLMmark \cite{Liu2023AAAI}}&18.748&11.056&0.009&7.22&43.50&	98.81&	99.92&		99.85\\
		\textbf{\multirow{6}{*}{DAIRstega}} &$\beta$=1, $\alpha$=8& \textbf{7.686} & \textbf{0.006} &\underline{0.017*}&\underline{79.91*}&\textbf{65.76}&56.82& 67.25& 75.40\\
		& $\beta$=1, $\alpha$=32&\underline{8.184*}&\underline{0.492*}&0.022& 76.97 &65.22&\textbf{54.05}& \textbf{62.28}&71.97\\
		& $\beta$=1, $\alpha$=48&8.219 &0.527&0.020& 78.70& \underline{65.25*}&\underline{54.98*}&\underline{65.51*}& \underline{71.40*}\\
		& $\beta$=0.5, $\alpha$=8&13.775& 6.082&0.018&\textbf{80.83}& 62.49&63.13& 65.67&\textbf{69.04}\\
		& $\beta$=0.5, $\alpha$=32&20.876 &13.184&0.029& 75.58& 60.89&68.73& 71.71 &76.41\\
		& $\beta$=0.5, $\alpha$=48&22.069& 14.377&\textbf{0.006}&72.40&60.29&69.48& 70.64& 79.46\\
		\bottomrule[1.2pt]
	\end{tabular}%
\end{table*}%

According to the comparison results in Table \ref{com_watermark}, the watermark texts generated by DAIR have excellent quality and consistency, and it is difficult for unauthorized receivers to detect the existence of the watermark text, avoiding the problem of watermark abuse. Only authorized receivers with the same model settings can detect and fully extract the watermark information.

\subsection{Longer stegos by DAIRstega}\label{sec35}

\begin{table*}[!htbp]
	\centering
	\footnotesize
	\setlength{\tabcolsep}{1.2mm}
	\caption{Examples of long stegos generated by DAIRstega. Bitstream is the binary of the secret. Secret can be not only texts, but also other modalities. In addition, we embed the image information. For details, see Table \ref{cm} in Appendix \ref{Appendix_E}.}\label{tab5}%
	\begin{tabular}{p{5.8em}p{6.5em}p{43em}p{2em}p{2.5em}}
		\toprule[1.2pt]
		\multicolumn{1}{p{6.5em}}{\footnotesize{Secret\&Bitsream}} & \multicolumn{1}{p{5em}}{\footnotesize{Instruction}} & \multicolumn{1}{c}{\footnotesize{Stegos}} & \multicolumn{1}{p{2em}}{\footnotesize{PPL}} & \multicolumn{1}{p{3em}}{\footnotesize{$\Delta$LDA}} \\
		\midrule[0.5pt]
		
		\multicolumn{1}{p{6.5em}}{\multirow{3}{=}{\newline{}\newline{}\newline{}\textbf{Secret}:\newline{} \underline{Love and peace}\newline{}\newline{} \textbf{Bitstream}:\newline{} 010011000110 111101110110 011001010010 000001100'00 101101110011 001000010000 001110000011 00101'011000 010110001101 100101 (114bit)}} & \multicolumn{1}{p{5em}}{\multirow{1}{=}{\newline{}\newline{}\newline{} Discuss the causes of the Great Depression.}} & The Great Depression was a severe economic downturn that lasted from 1929 to 1939. It was the worst economic depression in the history of the United States and it affected many countries around the world. The causes of the Great Depression are numerous and complex, but some of the main factors include overproduction, the bursting of the stock market bubble, banking failures, and government policies. One of the primary causes of the Great Depression was overproduction. The United States economy was booming in the 1920s, and many businesses and industries expanded their production to meet the growing demand. However, this resulted in an oversupply of goods and services, which led to a decrease in prices and a decrease in demand. This decrease in demand caused a drop in profits, which led to businesses cutting back on production and laying off workers. & \multicolumn{1}{p{2em}}{\multirow{1}{=}{\newline{}\newline{}\newline{}\newline{} 2.145}} & \multicolumn{1}{p{2em}}{\multirow{1}{=}{\newline{}\newline{}\newline{}\newline{} 0.004}} \\
		\cline{2-5}          & Explain the use of word embeddings in Natural Language Processing. & Word embeddings are a form of representation learning in natural language processing (NLP). Word embeddings are vectors of real numbers that represent the meaning of words. These vectors are learned from large amounts of text data and can be used to model the relationship between words and their contexts. Word embeddings can be used in various NLP tasks such as sentiment analysis, machine translation, and text classification. & \multicolumn{1}{p{2em}}{\multirow{1}{=}{\newline{}\newline{}3.399}}   & \multicolumn{1}{p{2em}}{\multirow{1}{=}{\newline{}\newline{} 0.006}} \\
		\cline{2-5}          & Give some strategies for getting over a fear of public speaking. & Public speaking is a common fear for many people, and it can be difficult to overcome. Here are some strategies that can help: 1. Practice: The more you practice public speaking, the more comfortable you will become. Start by practicing in front of a small group of friends or family members, and gradually work your way up to larger audiences. 2. Visualize: Before giving a speech, visualize yourself giving a successful presentation. Imagine yourself confidently delivering your speech, and see yourself receiving positive feedback from the audience. 3. Breathe: Take deep, slow breaths before giving a speech. & \multicolumn{1}{p{2em}}{\multirow{1}{=}{\newline{}\newline{}\newline{} 2.046}} & \multicolumn{1}{p{2em}}{\multirow{1}{=}{\newline{}\newline{}\newline{} 0.007}} \\
		\bottomrule[1.2pt]
	\end{tabular}%
\end{table*}%

The secret will change the next token. This makes it difficult to maintain logical coherence when generating longer stegos. So, longer stegos with high quality is a problem, and there is little research in this task. Fortunately, we found that DAIRstega also has this ability. Examples of embedded text and cross-modal secrets are shown in Table \ref{tab5} and Table \ref{cm}.

\subsection{Ablation of the different LLMs}\label{sec36}
We performed ablation experiments on different LLMs, and the results are shown in Table \ref{ae}.

\begin{table}[!htbp]
	\centering
	\scriptsize
	\setlength{\tabcolsep}{1.3mm}
	\caption{Ablation on different LLMs. \textbf{Bold} is the best result. Here we only take ``$\beta$ = 1, $\alpha$ = 32" as an example.}\label{ae}%
	\begin{tabular}{c||cccccc}
		\toprule[1.2pt]
		\multirow{2}{*}{$\beta$=1, $\alpha$=32} & \multicolumn{1}{c|}{Perceptual}&\multicolumn{5}{c}{Statistical}\\
		&\multicolumn{1}{c|}{PPL $\downarrow$}  & CS $\uparrow$    & JSD $\downarrow$   & ED $\downarrow$   &MD $\downarrow$    & $\Delta$DP $\downarrow$  \\
		\midrule[0.5pt]
		LLaMA2-7B & \multicolumn{1}{c|}{\textbf{8.184}} & \textbf{97.65} & \textbf{38.48} & \textbf{0.217} &\textbf{6.293} & \textbf{2.35} \\
		LLaMA2-13B & \multicolumn{1}{c|}{8.431} & 96.32 & 43.93 & 0.271 &8.034 & 3.68\\
		\midrule[1.2pt]
		
		\multirow{2}{*}{$\beta$=1, $\alpha$=32}&\multicolumn{4}{c|}{Semantic}&\multicolumn{2}{c}{Anti-steganalysis} \\
		& Mauve $\uparrow$ & BLEU $\uparrow$ &$\Delta$LDA $\downarrow$ & \multicolumn{1}{c|}{Score $\uparrow$}  &TS\_CSW  $\downarrow$ & UP4LS $\downarrow$\\
		\midrule[0.5pt]
		LLaMA2-7B & 76.97 & \textbf{77.85} &0.022 & \multicolumn{1}{c|}{\textbf{65.22}}&  \textbf{54.05} & 71.97\\
		LLaMA2-13B& \textbf{88.38} & 72.31&\textbf{0.019} & \multicolumn{1}{c|}{61.66}& 54.73 & \textbf{71.12} \\
		\bottomrule[1.2pt]
	\end{tabular}%
\end{table}%

According to Table \ref{ae}, it is not possible to generate stegos with stronger concealment by using larger LLMs. This is because steganography needs to embed secrets. The stego quality is more relevant with the embedding way rather than using larger LLMs.

\section{Conclusion}\label{sec4}

To enhance the concealment of stegos, this paper proposes a steganography scheme based on allocated roulette area, namely DAIRstega. It uses the idea of the non-uniform roulette wheel and takes the CP of tokens as an important basis for allocating the number of codings. Then, the allocation function and three constraints are designed to optimize the allocation process. Experiments show that the proposed embedding way and DAIRstega have achieved excellent performance in various concealment and anti-steganalysis. Last but not least, to show the proposed scheme concisely, we did not conduct the results of all allocation functions and LLMs.

\section{Outlook}\label{sec5}
By reflecting on this field, we found gaps in it that can be studied by scholars. \textbf{1.} This paper preliminarily explores the performance of the proposed method in long stegos. However, the research on long stegos is still in the basic stage. So, given the limited studies on longer and more diverse steganography, the scheme can be designed to improve the length and diversity. \textbf{2.} There is currently a lack of high-quality steganography schemes for closed-source LLMs. So, knowledge Graphs or Chain of Thought can enhance the prompts, while encoding text features allows for convenient and completely concealed steganography. \textbf{3.} The scheme in this paper can also be used as a watermarking for LLMs, which can make up for the insufficiency of the watermarking for LLMs being difficult to completely extract. Therefore, a new watermarking for LLMs can be designed by referring to the scheme in this paper. \textbf{4.} At present, linguistic steganography can achieve strong concealment and anti-steganalysis capabilities, that is, it is relatively safe. However, the sender of the stego is dangerous. Therefore, virtual social robots can be developed that utilize secrets to generate multi-turn conversations, promising contactless, covert communication that ensures the sender's safety. \textbf{5.} Finally, we found that existing steganalysis methods have poor detection performance on DAIRstega's stegos. Therefore, it is essential to advance steganalysis techniques.

\section*{Acknowledgment}\label{sec6}
We would like to express our gratitude to the MindSpore Community for their invaluable support and resources throughout this research. Thanks for the support provided by MindSpore Community. Thank the editor and reviewers of ``\textit{Applied Soft Computing}" for their review and constructive suggestions.

\section*{Declaration of competing interest}\label{sec8}
The authors declare no conflict of interest.

\section*{Funding}\label{sec9}
This work is supported by the National Natural Science Foundation of China (Grant U21B2020) and supported by BUPT Excellent Ph.D. Students Foundation (Grant CX2023120, CX20241055).

\section*{Author contributions}\label{sec10}
Yihao Wang: Conceptualization; Data curation; Investigation; Methodology; Software; Funding acquisition; original draft. Ruiqi Song: Data curation; Investigation; Resources; Software; Validation; original draft. Lingxiao Li: Data curation; Resources; Software; Funding acquisition; original draft; review \& editing. \textit{\textbf{Yihao Wang and Lingxiao Li contributed equally to this work and should be considered co-first authors.}} Ru Zhang: Funding acquisition; Investigation; Project administration; Supervision; review \& editing. Jianyi Liu: Formal analysis; Supervision; Investigation; Validation; review \& editing. 

\section*{Ethical Statement}\label{sec11}
This research does not involve human participants, personal data, or animal subjects; therefore, it does not raise any direct ethical concerns. The data used in this study were sourced from publicly available datasets or generated by applicable laws and regulations.

\newpage
\bibliographystyle{elsarticle-num}
\bibliography{references}

\begin{thebibliography}{10}
\expandafter\ifx\csname url\endcsname\relax
  \def\url#1{\texttt{#1}}\fi
\expandafter\ifx\csname urlprefix\endcsname\relax\def\urlprefix{URL }\fi
\expandafter\ifx\csname href\endcsname\relax
  \def\href#1#2{#2} \def\path#1{#1}\fi

\bibitem{Shannon}
C.~E. Shannon, Communication theory of secrecy systems, The Bell system
  technical journal 28~(4) (1949) 656--715.

\bibitem{Yang2023TDSC}
T.~Yang, H.~Wu, B.~Yi, G.~Feng, X.~Zhang, Semantic-preserving linguistic
  steganography by pivot translation and semantic-aware bins coding, IEEE
  Transactions on Dependable and Secure Computing (2023) 139--152.

\bibitem{VAE-Stega}
Z.~Yang, S.~Zhang, Y.~Hu, Z.~Hu, Y.~Huang, Vae-stega: linguistic steganography
  based on variational auto-encoder, IEEE Transactions on Information Forensics
  and Security 16 (2021) 880--895.

\bibitem{Lu2023IJCNN}
T.~Lu, G.~Liu, R.~Zhang, T.~Ju, Neural linguistic steganography with
  controllable security, in: Proceeding of the Joint Conference on Neural
  Networks, 2023, pp. 1--8.

\bibitem{Hi-Stega}
H.~Wang, Z.~Yang, J.~Yang, Y.~Gao, Y.~Huang, Hi-stega: A hierarchical
  linguistic steganography framework combining retrieval and generation, in:
  Proceeding of International Conference on Neural Information Processing
  (ICONIP), 2023, pp. 41--54.

\bibitem{Huo2016ICCC}
L.~Huo, Y.~Xiao, Synonym substitution-based steganographic algorithm with
  vector distance of two-gram dependency collocations, in: Proceeding of
  International Conference on Computer and Communications, 2016, pp.
  2776--2780.

\bibitem{Kim2010}
M.~Kim, O.~Zaiane, R.~Goebel, Natural language watermarking based on syntactic
  displacement and morphological division, in: Proceeding of the Annual IEEE
  Computer Software and Applications Conference, 2010.

\bibitem{Topic}
Y.~Li, J.~Zhang, Z.~Yang, R.~Zhang, Topic-aware neural linguistic steganography
  based on knowledge graphs, ACM Transactions on Data Science (2021) 1--13.

\bibitem{RNN-Stega}
Z.~Yang, X.~Guo, Z.~Chen, Y.~Huang, Y.~Zhang, Rnn-stega: Linguistic
  steganography based on recurrent neural networks, IEEE Transactions on
  Information Forensics and Security 14~(5) (2019) 1280--1295.

\bibitem{Ding2023TCDS}
C.~Ding, Z.~Fu, Q.~Yu, F.~Wang, X.~Chen, Joint linguistic steganography with
  bert masked language model and graph attention network, IEEE Transactions on
  Cognitive and Developmental Systems (2023) 1\href
  {https://doi.org/10.1109/TCDS.2023.3296413}
  {\path{doi:10.1109/TCDS.2023.3296413}}.

\bibitem{GAN-APD}
X.~Zhou, W.~Peng, B.~Yang, J.~Wen, Y.~Xue, P.~Zhong, Linguistic steganography
  based on adaptive probability distribution, IEEE Transactions on Dependable
  and Secure Computing 19~(5) (2021) 2982--2997.

\bibitem{ADG}
S.~Zhang, Z.~Yang, J.~Yang, Y.~Huang, Provably secure generative linguistic
  steganography, in: Proceeding of the International Joint Conference on
  Natural Language Processing (ACL-IJCNLP), 2021, pp. 3046--3055.

\bibitem{Meteor}
G.~Kaptchuk, T.~M. Jois, M.~Green, A.~D. Rubin, Meteor: Cryptographically
  secure steganography for realistic distributions, in: Proceedings of the 2021
  ACM SIGSAC Conference on Computer and Communications Security, 2021, pp.
  1529--1548.

\bibitem{FEFT}
Z.~Yang, Y.~Huang, Y.~Zhang, A fast and efficient text steganalysis method,
  IEEE Signal Processing Letters 26~(4) (2019) 627--631.

\bibitem{tmode}
Y.~Tang, Y.~Wang, R.~Zhang, J.~Liu, Linguistic steganalysis via llms: Two modes
  for efficient detection of strongly concealed stego., IEEE Signal Processing
  Letters 32 (2024) 541--545.

\bibitem{TinaFang}
T.~Fang, M.~Jaggi, K.~Argyraki, Generating steganographic text with lstms, in:
  Proceedings of the 55th Annual Meeting of the Association for Computational
  Linguistics (ACL)- Student Research Workshop, 2017, pp. 100--106.

\bibitem{Discop}
J.~Ding, K.~Chen, Y.~Wang, N.~Zhao, W.~Zhang, N.~Yu, Discop: Provably secure
  steganography in practice based on distribution copies, in: IEEE Symposium on
  Security and Privacy (SP), 2023, pp. 2238--2255.

\bibitem{GPT4}
OpenAI, Gpt-4 technical report, arXiv preprint (2023).
\newblock \href {http://arxiv.org/abs/arXiv:2303.08774}
  {\path{arXiv:arXiv:2303.08774}}.

\bibitem{G1}
A.~Alsajri, H.~A. Salman, A.~Steiti, Generative models in natural language
  processing: A comparative study of chatgpt and gemini, Babylonian Journal of
  Artificial Intelligence 2024 (2024) 134--145.

\bibitem{G3}
bdulazeez Alsajri, M.~Aljanabi, The impact of chatgpt on social media, MEDAAD
  2024 (2024) 9--14.

\bibitem{LLaMA2}
Meta, Llama 2: Open foundation and fine-tuned chat models, arXiv preprint
  (2023).
\newblock \href {http://arxiv.org/abs/arXiv:2307.09288}
  {\path{arXiv:arXiv:2307.09288}}.

\bibitem{Liu2023AAAI}
P.~Liu, P.~Cheng, F.~Li, W.~Du, H.~Zhao, G.~Liu, Plmmark: a secure and robust
  black-box watermarking framework for pre-trained language models., in:
  Proceedings of the Thirty-Seventh AAAI Conference on Artificial Intelligence,
  2023, pp. 14991--14999.

\bibitem{Wu2023ICML}
X.~Wu, C.~Li, R.~Y. Aminabadi, Z.~Yao, Y.~He, Understanding int4 quantization
  for language models: latency speedup, composability, and failure cases, in:
  Proceedings of International Conference on Machine Learning (ICML), 2023, pp.
  37524--37539.

\bibitem{Cai2021ICML}
X.~Cai, J.~Huang, Y.~Bian, K.~Church, Isotropy in the contextual embedding
  space: Clusters and manifolds, in: Proceedings of International Conference on
  Machine Learning (ICML), 2021.

\bibitem{MAUVE}
K.~Pillutla, S.~Swayamdipta, R.~Zellers, J.~Thickstun, S.~Welleck, Mauve:
  Measuring the gap between neural text and human text using divergence
  frontiers, in: Proceedings of the 35th Conference on Neural Information
  Processing Systems (NeurIPS), 2021, pp. 4816--4828.

\bibitem{BLEU}
K.~Papineni, S.~Roukos, T.~Ward, W.-J. Zhu, Bleu: a method for automatic
  evaluation of machine translation, in: Proceedings of the 40th annual meeting
  of the Association for Computational Linguistics, 2002, pp. 311--318.

\bibitem{bertscore}
T.~Zhang, V.~Kishore, F.~Wu, K.~Q. Weinberger, Y.~Artzi, Bertscore: Evaluating
  text generation with bert, in: Proceedings of the 35th Conference on Neural
  Information Processing Systems (NeurIPS), 2020.

\bibitem{G2}
A.~Alsajri, A.~Steiti, Intrusion detection system based on machine learning
  algorithms: (svm and genetic algorithm), Babylonian Journal of Artificial
  Intelligence 2024 (2024) 15--29.

\bibitem{RLS-DTS}
Y.~Wang, R.~Zhang, J.~Liu, Rls-dts: Reinforcement-learning linguistic
  steganalysis in distribution-transformed scenario, IEEE Signal Processing
  Letters 30 (2023) 1232--1236.

\bibitem{LS_CNN}
J.~Wen, X.~Zhou, P.~Zhong, Y.~Xue, Convolutional neural network based text
  steganalysis, IEEE Signal Processing Letters 26~(3) (2019) 460--464.

\bibitem{TS_CSW}
Z.~Yang, Y.~Huang, Y.~Zhang, Ts-csw: text steganalysis and hidden capacity
  estimation based on convolutional sliding windows, Multimedia Tools and
  Applications 79 (2020) 18293--18316.

\bibitem{EILG}
Q.~Xu, R.~Zhang, J.~Liu, Linguistic steganalysis by enhancing and integrating
  local and global features, IEEE Signal Processing Letters 30 (2023) 16--20.

\bibitem{UP4LS}
Y.~Wang, R.~Song, L.~Li, Y.~Tang, R.~Zhang, J.~Liu, Up4ls: User profile
  constructed by multiple attributes for enhancing linguistic steganalysis,
  arXiv preprint (2023).
\newblock \href {http://arxiv.org/abs/arXiv:2311.01775}
  {\path{arXiv:arXiv:2311.01775}}.

\bibitem{rbilstmc}
Y.~Niu, J.~Wen, P.~Zhong, Y.~Xue, A hybrid r-bilstm-c neural network based text
  steganalysis., IEEE Signal Processing Letters 26~(12) (2019) 107--1911.

\bibitem{Zou}
J.~Zou, Z.~Yang, S.~Zhang, S.~Rehman, Y.~Huang, High-performance linguistic
  steganalysis, capacity estimation and steganographic positioning., in:
  Proceeding of the International Workshop on Digital Watermarking (IWDW),
  2021, pp. 80--93.

\bibitem{SSLS}
Y.~Xu, T.~Zhao, P.~Zhong, Small-scale linguistic steganalysis for
  multi-concealed scenarios, IEEE Signal Processing Letters 29 (2022) 130--134.

\bibitem{sesy}
J.~Yang, Z.~Yang, S.~Zhang, H.~Tu, Y.~Huang, Sesy: Linguistic steganalysis
  framework integrating semantic and syntactic features., IEEE Signal
  Processing Letters 29 (2021) 31--35.

\bibitem{llmsteganalysis}
M.~Bai, J.~Yang, K.~Pang, H.~Wang, Y.~Huang, Towards next-generation
  steganalysis: Llms unleash the power of detecting steganography, arXiv
  preprint (2024).
\newblock \href {http://arxiv.org/abs/arXiv:2405.09090}
  {\path{arXiv:arXiv:2405.09090}}.

\bibitem{watermark1}
J.~Kirchenbauer, J.~Geiping, Y.~Wen, J.~Katz, I.~Miers, T.~Goldstein, A
  watermark for large language models, in: Proceedings of International
  Conference on Machine Learning (ICML), 2023.

\bibitem{watermark2}
R.~Zhang, S.~S. Hussain, P.~Neekhara, F.~Koushanfar, Remark-llm: A robust and
  efficient watermarking framework for generative large language models, in:
  Proceedings of Usenix security, 2024.

\end{thebibliography}


\newpage
\onecolumn
\section{Appendix} 
\label{appendix}
\subsection{Supplement the effect of the proposed embedding way}\label{Appendix_B}

This section gives the detailed values of Table 1, including the performance comparison of different embedding ways under different parameters and corresponding Bpw. The specific comparison results on Text quality, Statistical analysis, and Discourse matching are shown in Table \ref{app_embedding1}, and the specific comparison results on anti-steganalysis are shown in Table \ref{app_embedding2}.

\begin{table}[!htbp]
	\centering
	\small
	\setlength{\tabcolsep}{0.42mm}
	\caption{\underline{Text quality, statistical analysis, and discourse-matching} comparison of stegos generated by DAIRstega and baselines. The embedding rates of each way include 1 Bpw (bit per word) and 2 Bpw. ``Avg." represents the overall performance. ``Cap'' represents the candidate pool. The meanings of ``bit" are the same in Table 2. \textbf{Bold} is the best overall result. ``\underline{ *}" is the suboptimal overall result. ``$\uparrow$" is that the higher the value, the better the result. ``$\downarrow$" means the lower the value, the better the result. For a detailed description of the embedding ways, please see section \ref{sec31}.}\label{app_embedding1}%
	\begin{tabular}{cc||cc|ccccc|cccc}
		\toprule[1.2pt]
		\multicolumn{1}{c}{\multirow{2}{*}{\textbf{Embedding}}} &\multicolumn{1}{c||}{\multirow{2}{*}{Param / Bpw}}&\multicolumn{2}{c|}{Text quality}&\multicolumn{5}{c|}{Statistical analysis}&\multicolumn{4}{c}{Discourse-matching}\\
		&& \multicolumn{1}{c}{PPL $\downarrow$} & \multicolumn{1}{c|}{$\Delta$Pcs $\downarrow$} & \multicolumn{1}{c}{CS $\uparrow$} & \multicolumn{1}{c}{JSD $\downarrow$} & \multicolumn{1}{c}{ED $\downarrow$ } & \multicolumn{1}{c}{MD $\downarrow$} & \multicolumn{1}{c|}{$\Delta$DP $\downarrow$} & \multicolumn{1}{c}{$\Delta$LDA $\downarrow$} & Mauve $\uparrow$ & BLEU $\uparrow$ & Score $\uparrow$ \\
		\midrule[0.5pt]
		
		\multirow{3}{*}{LLMs+HC} & bit = 1 / 1.00 & 13.331  & 5.639  & 96.54  & 45.79  & 0.263  & 8.255  & 3.46  & 0.014  & 70.60\textsubscript{$\pm$27.36} & 54.26\textsubscript{$\pm$4.25} & 58.44\textsubscript{$\pm$4.07} \\
		& bit = 2 / 2.52 & 24.334  & 16.641  & 90.57  & 47.35  & 0.350  & 9.354  & 9.43  & 0.009  & 64.88\textsubscript{$\pm$30.03} & 55.00\textsubscript{$\pm$5.77} & 56.19\textsubscript{$\pm$4.01} \\
		\rowcolor[rgb]{ .900,  .900,  .900}\cellcolor{blue!0}& Avg. & 18.832  & 11.140  & 93.55  & 46.57  & 0.307  & 8.805  & 6.45  & \textbf{0.011}  & 67.74  & 54.63  & 57.32  \\
		\midrule[0.5pt]
		
		\multirow{3}{*}{LLMs+ADG} & $\tau$ = 0.3 / 1.34 & 12.750  & 5.058 
		& 97.20  & 45.59  & 0.240  & 8.316  & 2.80  & 0.010  & 77.86\textsubscript{$\pm$23.65} & 71.79\textsubscript{$\pm$3.03} & 59.18\textsubscript{$\pm$5.84} \\
		& $\tau$ = 0.65 / 2.99 & 16.175  & 8.483  & 95.95  & 44.74  & 0.331  & 8.478  & 4.05  & 0.018  & 75.60\textsubscript{$\pm$27.02} & 62.60\textsubscript{$\pm$4.68} & 58.60\textsubscript{$\pm$5.69} \\
		& Avg. & \underline{14.463*}  & \underline{6.770*}  & \textbf{96.58}  & \underline{45.17*}  & 0.285  & \underline{8.397*}  & \textbf{3.42}  & 0.014  & \underline{76.73*}  & \underline{67.20*}  & \underline{58.35*}  \\
		\midrule[0.5pt]
		
		\multirow{3}{*}{LLMs+AC} & Cap = 8 / 1.25 & 73.454 & 65.762  & 82.16 & 51.64 & 0.746 & 15.415 & 17.84 & 0.014 & 35.64\textsubscript{$\pm$22.65} & 50.92\textsubscript{$\pm$29.30} & 45.10\textsubscript{$\pm$3.07} \\
		& Cap = 32 / 1.32 & 86.238 & 78.545 & 80.38 & 51.21 & 0.829 & 14.194 & 19.62 & 0.012 & 37.18\textsubscript{$\pm$24.59} & 48.01\textsubscript{$\pm$27.45} & 47.51\textsubscript{$\pm$2.76} \\
		\rowcolor[rgb]{ .900,  .900,  .900}\cellcolor{blue!0}& Avg. & 79.846 & 72.154 & 81.27 & 51.43 & 0.788 & 14.805 & 18.73 & \underline{0.013*} & 36.41  & 49.47  & 46.31  \\
		\midrule[0.5pt]
		
		\multirow{3}{*}{LLMs+FLC} & bit = 1 / 1.34 & 12.883  & 5.190  & 97.04  & 46.56  & 0.241  & 8.280  & 2.96  & 0.031  & 79.48\textsubscript{$\pm$26.53} & 54.88\textsubscript{$\pm$6.33} & 58.44\textsubscript{$\pm$4.64} \\
		& bit = 2 / 2.67 & 30.481  & 22.788 & 93.86 & 46.30  & 0.328 & 8.939 & 6.14  & 0.011 & 61.10\textsubscript{$\pm$29.30} & 58.21\textsubscript{$\pm$9.88} & 55.20\textsubscript{$\pm$3.39} \\
		\rowcolor[rgb]{ .900,  .900,  .900}\cellcolor{blue!0}& Avg. & 21.682  & 13.989  & 95.45  & 46.43  & \underline{0.284*}  & 8.610  & 4.55  & 0.021  & 70.29  & 56.55  & 56.82  \\
		\midrule[0.5pt]
		
		\multirow{9}{*}{LLMs+\textbf{Ours}} & $\beta$=1 $\alpha$=8 / 1.10 & 7.686  & 0.006  & 97.20  & 38.86  & 0.237  & 6.458  & 2.80  & 0.017  & 79.91\textsubscript{$\pm$26.75} & 77.88\textsubscript{$\pm$30.40} & 65.76\textsubscript{$\pm$4.13} \\
		& $\beta$=1 $\alpha$=32 / 1.11 & 8.184  & 0.492  & 97.65  & 38.48  & 0.217  & 6.293  & 2.35  & 0.022  & 76.97\textsubscript{$\pm$26.47} & 77.85\textsubscript{$\pm$30.42} & 65.22\textsubscript{$\pm$4.61} \\
		& $\beta$=1 $\alpha$=48 / 1.13 & 8.219  & 0.527  & 97.45  & 38.46  & 0.226  & 6.273  & 2.55  & 0.020  & 78.70\textsubscript{$\pm$25.51} & 76.53\textsubscript{$\pm$31.71} & 65.25\textsubscript{$\pm$4.78} \\
		&Avg. (1Bpw)&8.030 &	0.342& 	97.43& 	38.60 &	0.227 &	6.341 &	2.57 &	0.020 &	78.53&	77.42	&65.41\\
		\cdashline{2-13}[3pt/2pt]
		
		& $\beta$=0.5 $\alpha$=8 / 1.89 & 13.775  & 6.082  & 97.39  & 37.80  & 0.229  & 7.204  & 2.61  & 0.018  & 80.83\textsubscript{$\pm$25.15} & 73.19\textsubscript{$\pm$26.93} & 62.49\textsubscript{$\pm$4.48} \\
		& $\beta$=0.5 $\alpha$=32 / 2.56 & 20.876  & 13.184  & 92.43  & 39.69  & 0.389  & 8.976  & 7.57  & 0.029  & 75.58\textsubscript{$\pm$20.11} & 66.99\textsubscript{$\pm$27.08} & 60.89\textsubscript{$\pm$3.85} \\
		& $\beta$=0.5 $\alpha$=48 / 2.58 & 22.069  & 14.377  & 92.44  & 38.31  & 0.389  & 8.592  & 7.56  & 0.006  & 72.40\textsubscript{$\pm$26.33} & 79.64\textsubscript{$\pm$12.47} & 60.29\textsubscript{$\pm$4.00} \\
		&Avg. (2Bpw)&18.907 &	11.214& 	94.09 &	38.60 &	0.336 &	8.258& 	5.91& 	0.018 &	76.27&	73.27&	61.22\\
		\cdashline{2-13}[3pt/2pt]
		
		\rowcolor[rgb]{ .900,  .900,  .900}\cellcolor{blue!0}& Avg. & \textbf{13.468}  & \textbf{5.778}  & \underline{95.76*}  & \textbf{38.60}  & \textbf{0.281}  & \textbf{7.299}  & \underline{4.24*}  & 0.019  & \textbf{77.40}  & \textbf{75.35}  & \textbf{63.32}  \\
		\bottomrule[1.2pt]
	\end{tabular}%
\begin{tablenotes}
	\item[] \footnotesize 1. HC (Huffman coding) is used in VAE-Stega \cite{VAE-Stega}, LSCS \cite{Lu2023IJCNN}, RNNstega \cite{RNN-Stega}, GAN-APD \cite{GAN-APD}, JLS \cite{Ding2023TCDS}, and Discop \cite{Discop}.
	\item[] 2. ADG (Adaptive dynamic grouping) is used in Hi-Stega \cite{Hi-Stega} and ADG \cite{ADG}.
	\item[] 3. AC (Arithmetic coding) is used in VAE-Stega \cite{VAE-Stega}, Hi-Stega \cite{Hi-Stega}, and Meteor \cite{Meteor}.
	\item[] 4. FLC (Fixed length coding) is used in RNNstega \cite{RNN-Stega} and JLS \cite{Ding2023TCDS}.
\end{tablenotes}
\end{table}%

\begin{table}[!htbp]
	\centering
	\small
	\setlength{\tabcolsep}{1.12mm}
	\caption{\underline{Anti-steganalysis} comparison of stegos generated by DAIRstega and baselines. The embedding rates of each way include 1Bpw (bit per word) and 2Bpw. ``Avg." represents the overall performance. \textbf{Bold} is the best overall result. ``\underline{ *}" is the suboptimal overall result. ``Cap'' represents the candidate pool. ``$\uparrow$" is that the higher the value, the better the result. ``$\downarrow$" means the lower the value, the better the result. For a detailed description of the embedding ways, please see section \ref{sec31}.}\label{app_embedding2}%
	\begin{tabular}{cc||cccccccc}
		\toprule[1.2pt]
		\multirow{2}{*}{\textbf{Embedding}} & \multirow{2}{*}{Param / Bpw} & \multicolumn{2}{c}{LS\_CNN \cite{LS_CNN}} & \multicolumn{2}{c}{TS\_CSW \cite{TS_CSW}} & \multicolumn{2}{c}{EILG \cite{EILG}} & \multicolumn{2}{c}{UP4LS \cite{UP4LS}} \\
		&&Acc $\downarrow$&F1 $\downarrow$&Acc $\downarrow$&F1 $\downarrow$&Acc $\downarrow$&F1 $\downarrow$&Acc $\downarrow$&F1 $\downarrow$\\
		\midrule
		\multirow{3}{*}{LLMs+HC} & bit = 1 / 1.00 & 77.62\textsubscript{$\pm$1.38} & 78.41\textsubscript{$\pm$1.65} & 78.09\textsubscript{$\pm$1.33} & 78.25\textsubscript{$\pm$1.21} & 78.58\textsubscript{$\pm$1.69} & 78.29\textsubscript{$\pm$2.24} & 86.73\textsubscript{$\pm$1.32} & 86.13\textsubscript{$\pm$0.50} \\
		& bit = 2 / 2.52 & 77.54\textsubscript{$\pm$1.50} & 77.57\textsubscript{$\pm$2.05} & 76.03\textsubscript{$\pm$1.74} & 76.09\textsubscript{$\pm$1.33} & 78.29\textsubscript{$\pm$0.72} & 78.95\textsubscript{$\pm$1.47} & 88.19\textsubscript{$\pm$1.00} & 88.43\textsubscript{$\pm$2.61} \\
		\rowcolor[rgb]{ .900,  .900,  .900}\cellcolor{blue!0}& Avg. & 77.58 & 77.99 & 77.06 & 77.17 & 78.44 & 78.62 & 87.46 & 87.28 \\
		\midrule
		\multirow{3}{*}{LLMs+ADG} & $\tau$ = 0.3 / 1.34 & 70.10\textsubscript{$\pm$0.66} & 70.17\textsubscript{$\pm$0.89} & 69.45\textsubscript{$\pm$2.05} & 69.34\textsubscript{$\pm$2.53} & 70.99\textsubscript{$\pm$3.15} & 70.49\textsubscript{$\pm$1.11} & 84.00\textsubscript{$\pm$0.83} & 84.23\textsubscript{$\pm$1.07} \\
		& $\tau$ = 0.65 / 2.99 & 63.06\textsubscript{$\pm$2.49} & 63.69\textsubscript{$\pm$3.61} & 62.22\textsubscript{$\pm$1.46} & 61.86\textsubscript{$\pm$1.18} & 66.14\textsubscript{$\pm$1.38} & 66.15\textsubscript{$\pm$1.38} & 80.42\textsubscript{$\pm$0.59} & 80.57\textsubscript{$\pm$0.58} \\
		\rowcolor[rgb]{ .900,  .900,  .900}\cellcolor{blue!0}& Avg. & \underline{66.58*} & \underline{66.93*} & \underline{65.84*} & \underline{65.60*} & \underline{68.57*} & \underline{68.32*} & \underline{82.21*} & \underline{82.40*} \\
		\midrule
		\multirow{3}{*}{LLMs+AC} & Cap = 8 / 1.25 & 90.15\textsubscript{$\pm$1.17} & 89.98\textsubscript{$\pm$1.21} & 89.69\textsubscript{$\pm$0.93} & 89.37\textsubscript{$\pm$0.88} & 86.92\textsubscript{$\pm$1.12} & 86.52\textsubscript{$\pm$0.96} & 93.62\textsubscript{$\pm$0.89} & 93.37\textsubscript{$\pm$0.40} \\
		& Cap = 32 / 1.32 & 88.45\textsubscript{$\pm$1.38} & 88.44\textsubscript{$\pm$1.25} & 90.80\textsubscript{$\pm$1.97} & 90.52\textsubscript{$\pm$2.18} & 86.44\textsubscript{$\pm$2.47} & 85.86\textsubscript{$\pm$3.10} & 94.03\textsubscript{$\pm$0.64} & 94.16\textsubscript{$\pm$0.95} \\
		\rowcolor[rgb]{ .900,  .900,  .900}\cellcolor{blue!0}& Avg. & 89.30 & 89.21 & 90.25 & 89.95 & 86.68 & 86.19 & 93.83 & 93.77 \\
		\midrule
		\multirow{3}{*}{LLMs+FLC} & bit = 1 / 1.34 & 78.53\textsubscript{$\pm$1.20} & 80.05\textsubscript{$\pm$1.39} & 82.72\textsubscript{$\pm$1.27} & 83.32\textsubscript{$\pm$1.10} & 76.67\textsubscript{$\pm$0.85} & 76.15\textsubscript{$\pm$1.96} & 89.03\textsubscript{$\pm$0.25} & 89.20\textsubscript{$\pm$0.80} \\
		& bit = 2 / 2.67 & 76.49\textsubscript{$\pm$1.15} & 76.54\textsubscript{$\pm$1.13} & 77.08\textsubscript{$\pm$2.06} & 77.26\textsubscript{$\pm$1.75} & 76.95\textsubscript{$\pm$1.69} & 76.46\textsubscript{$\pm$1.74} & 90.62\textsubscript{$\pm$0.89} & 90.37\textsubscript{$\pm$0.40} \\
		\rowcolor[rgb]{ .900,  .900,  .900}\cellcolor{blue!0}& Avg. & 77.51  & 78.30  & 79.90  & 80.29  & 76.81  & 76.31  & 89.83  & 89.79  \\
		\midrule
		\multirow{9}{*}{\textbf{Ours}} & $\beta$=1 $\alpha$=8 / 1.10 & 59.50\textsubscript{$\pm$2.77} & 60.15\textsubscript{$\pm$10.59} & 56.82\textsubscript{$\pm$4.09} & 64.21\textsubscript{$\pm$10.03} & 67.25\textsubscript{$\pm$3.23} & 68.04\textsubscript{$\pm$1.53} & 75.40\textsubscript{$\pm$0.37} & 76.26\textsubscript{$\pm$1.28} \\
		& $\beta$=1 $\alpha$=32 / 1.11 & 59.35\textsubscript{$\pm$1.25} & 59.99\textsubscript{$\pm$5.51} & 54.05\textsubscript{$\pm$4.72} & 61.68\textsubscript{$\pm$11.17} & 62.28\textsubscript{$\pm$0.50} & 62.66\textsubscript{$\pm$2.31} & 71.97\textsubscript{$\pm$0.64} & 72.75\textsubscript{$\pm$1.21} \\
		& $\beta$=1 $\alpha$=48 / 1.13 & 58.00\textsubscript{$\pm$2.37} & 56.08\textsubscript{$\pm$13.79} & 54.98\textsubscript{$\pm$3.14} & 59.17\textsubscript{$\pm$11.37} & 65.51\textsubscript{$\pm$2.23} & 66.78\textsubscript{$\pm$3.91} & 71.40\textsubscript{$\pm$0.65} & 70.89\textsubscript{$\pm$2.05} \\
		&Avg. (1Bpw)&58.95 &	58.74& 	55.28 &	61.69 &	65.01 &	65.83 &	72.92 &	73.30 \\
		\cdashline{2-10}[3pt/2pt]
		
		& $\beta$=0.5 $\alpha$=8 / 1.89 & 62.88\textsubscript{$\pm$1.53} & 62.21\textsubscript{$\pm$1.83} & 63.13\textsubscript{$\pm$1.44} & 59.77\textsubscript{$\pm$5.36} & 65.67\textsubscript{$\pm$1.56} & 65.84\textsubscript{$\pm$2.20} & 69.04\textsubscript{$\pm$0.36} & 68.23\textsubscript{$\pm$2.50} \\
		& $\beta$=0.5 $\alpha$=32 / 2.56 & 72.06\textsubscript{$\pm$1.17} & 71.78\textsubscript{$\pm$1.21} & 68.73\textsubscript{$\pm$1.29} & 68.19\textsubscript{$\pm$3.44} & 71.71\textsubscript{$\pm$1.95} & 70.34\textsubscript{$\pm$2.41} & 76.41\textsubscript{$\pm$2.56} & 75.90\textsubscript{$\pm$2.73} \\
		& $\beta$=0.5 $\alpha$=48 / 2.58 & 70.17\textsubscript{$\pm$1.81} & 70.40\textsubscript{$\pm$3.11} & 69.48\textsubscript{$\pm$0.68} & 67.80\textsubscript{$\pm$1.98} & 70.64\textsubscript{$\pm$1.61} & 71.00\textsubscript{$\pm$2.23} & 79.46\textsubscript{$\pm$0.56} & 78.28\textsubscript{$\pm$1.13} \\
		&Avg. (2Bpw)&68.37&	68.13 	&67.11 &	65.25 &	69.34 &	69.06 &	74.97 &	74.14 \\
		\cdashline{2-10}[3pt/2pt]
		
		\rowcolor[rgb]{ .900,  .900,  .900}\cellcolor{blue!0}& Avg. & \textbf{63.66}  & \textbf{63.44}  & \textbf{61.20}  & \textbf{63.47}  & \textbf{67.18}  & \textbf{67.44}  & \textbf{73.95}  & \textbf{73.72}  \\
		
		\bottomrule[1.2pt]
	\end{tabular}%
\begin{tablenotes}
	\item[] \footnotesize 1. HC (Huffman coding) is used in VAE-Stega \cite{VAE-Stega}, LSCS \cite{Lu2023IJCNN}, RNNstega \cite{RNN-Stega}, GAN-APD \cite{GAN-APD}, JLS \cite{Ding2023TCDS}, and Discop \cite{Discop}.
	\item[] 2. ADG (Adaptive dynamic grouping) is used in Hi-Stega \cite{Hi-Stega} and ADG \cite{ADG}.
	\item[] 3. AC (Arithmetic coding) is used in VAE-Stega \cite{VAE-Stega}, Hi-Stega \cite{Hi-Stega}, and Meteor \cite{Meteor}.
	\item[] 4. FLC (Fixed length coding) is used in RNNstega \cite{RNN-Stega} and JLS \cite{Ding2023TCDS}.
\end{tablenotes}
\end{table}%

\FloatBarrier

\subsection{Complete data for Table \ref{com_dm} and Table \ref{com_as}}\label{Appendix_C}
In this section, we give the complete data of Table \ref{com_dm} and Table \ref{com_as}, as shown in Table \ref{app_t4} and Table \ref{app_t5}. Table \ref{app_t4} contains the standard deviations of many datasets with different discourse, and Table \ref{app_t5} contains the standard deviations of multiple runs.

\begin{table}[H]
	\centering
	\small
	\caption{Complete data for Table \ref{com_dm}. The meanings of ``bin" and ``bit" are the same in Table \ref{com_tq}. ``a\textsubscript{$\pm$b}" represents ``Avg.\textsubscript{$\pm$Std}". \textbf{Bold} is the best result. ``\underline{ *}" is the suboptimal result. The ``Std"s of DAIRstega Mauve and BLEU values are bigger. This is not due to the instability of 5-time runs, but is obtained from the effect on dozens of different discourse datasets, and represents the overall performance.}\label{app_t4}
	\begin{tabular}{ccc||ccc}
		\toprule[1.2pt]
		\textbf{Schemes}&\multicolumn{2}{c||}{Param / Bpw}&Mauve $\uparrow$&BLEU $\uparrow$&\multicolumn{1}{c}{Score $\uparrow$}\\
		\midrule[0.5pt]
		
		\multicolumn{1}{c}{\multirow{2}{*}{Fang \cite{TinaFang}}} & \multicolumn{2}{c||}{bin = 1 / 1.00}  &  2.15\textsubscript{$\pm$0.87} & 0.90\textsubscript{$\pm$1.58} & 42.58\textsubscript{$\pm$1.30} \\
		& \multicolumn{2}{c||}{bin = 3 / 3.00} & 1.96\textsubscript{$\pm$0.68} & 0.90\textsubscript{$\pm$1.59}&  41.65\textsubscript{$\pm$1.33}\\     
		\midrule[0.5pt]
		
		\multicolumn{1}{c}{\multirow{2}{*}{RNNstega \cite{RNN-Stega}}} & \multicolumn{2}{c||}{bit = 1 / 1.00} &   12.73\textsubscript{$\pm$10.56}  & 1.87\textsubscript{$\pm$3.28}&  52.57\textsubscript{$\pm$2.20}\\
		& \multicolumn{2}{c||}{bit = 3 / 2.63} &11.97\textsubscript{$\pm$10.56}  & 1.83\textsubscript{$\pm$3.21}&  52.36\textsubscript{$\pm$1.80}\\
		\midrule[0.5pt]
		
		\multirow{1}{*}{ADG \cite{ADG}} & \multicolumn{2}{c||}{$\tau$ = 0.5 / 4.38}& 18.54\textsubscript{$\pm$13.89}  & 2.72\textsubscript{$\pm$4.79}&  47.55\textsubscript{$\pm$2.25} \\
		\midrule[0.5pt]
		
		LSCS \cite{Lu2023IJCNN}&\multicolumn{2}{c||}{- / 1.12}& 2.40\textsubscript{$\pm$1.78}& 31.95\textsubscript{$\pm$12.89}&59.22\textsubscript{$\pm$3.14}\\
		\midrule[0.5pt]
		
		PLMmark \cite{Liu2023AAAI} &\multicolumn{2}{c||}{- / -}&	7.22\textsubscript{$\pm$4.88}&	0.15\textsubscript{$\pm$0.27}&		43.50\textsubscript{$\pm$0.79} \\
		\midrule[0.5pt]
		
		\multicolumn{1}{c}{\multirow{6}{*}{\textbf{DAIRstega}}} &\multirow{3}{*}{$\beta$ = 1}&$\alpha$ = 8 / 1.10& \underline{79.91\textsubscript{$\pm$26.75}*}  & 76.37\textsubscript{$\pm$31.85} & \textbf{65.76\textsubscript{$\pm$4.13}} \\
		&&$\alpha$ = 32 / 1.11&76.97\textsubscript{$\pm$26.47}  & \underline{77.85\textsubscript{$\pm$30.42}*} & 65.22\textsubscript{$\pm$4.61} \\
		&&$\alpha$ = 48 / 1.13 &78.70\textsubscript{$\pm$25.51}  & 76.53\textsubscript{$\pm$31.71} &  \underline{65.25\textsubscript{$\pm$4.78}*}\\
		\cdashline{2-6}[3pt/2pt]
		
		&\multirow{3}{*}{$\beta$ = 0.5}&$\alpha$ = 8 / 1.89&\textbf{80.83\textsubscript{$\pm$25.12}}&73.19\textsubscript{$\pm$26.93}&62.49\textsubscript{$\pm$4.48}\\
		&&$\alpha$ = 32 / 2.56&75.58\textsubscript{$\pm$20.11}&66.99\textsubscript{$\pm$27.08}&60.89\textsubscript{$\pm$3.85}\\
		&&$\alpha$ = 48 / 2.58&72.40\textsubscript{$\pm$26.33}&\textbf{79.64\textsubscript{$\pm$12.47}}&60.29\textsubscript{$\pm$4.00}\\
		\bottomrule[1.2pt]
	\end{tabular}%
\end{table}%

\begin{table*}[!htbp]
	\centering
	\small
	\setlength{\tabcolsep}{0.33mm}
	\caption{Complete data for Table \ref{com_as}. The meanings of ``bin" and ``bit"are the same in Table \ref{com_tq}. ``a\textsubscript{$\pm$b}" represents ``Avg.\textsubscript{$\pm$Std}". ``Avg" and ``Std" in this table are obtained by running 5 times. \textbf{Bold} is the best result. ``\underline{ *}" is the suboptimal result.}\label{app_t5}
	\begin{tabular}{ccc||ccccccccccc}
		\toprule[1.2pt]
		\multirow{2}{*}{\textbf{Schemes}}&\multicolumn{2}{c||}{\multirow{2}{*}{Param / Bpw}}& \multicolumn{2}{c}{LS\_CNN \cite{LS_CNN}} & \multicolumn{2}{c}{TS\_CSW \cite{TS_CSW}} & \multicolumn{2}{c}{EILG \cite{EILG}} & \multicolumn{2}{c}{UP4LS \cite{UP4LS}} \\
		&&& Acc $\downarrow$ & F1 $\downarrow$ & Acc $\downarrow$& F1 $\downarrow$ & Acc $\downarrow$& F1 $\downarrow$ & Acc $\downarrow$& F1 $\downarrow$\\
		\midrule[0.5pt]
		
		\multicolumn{1}{c}{\multirow{2}{*}{Fang \cite{TinaFang}}} & \multicolumn{2}{c||}{bin = 1 / 1.00}  &   99.70\textsubscript{$\pm$0.29} & 99.68\textsubscript{$\pm$0.31} & 88.34\textsubscript{$\pm$21.22} & 91.01\textsubscript{$\pm$15.69} & 99.50\textsubscript{$\pm$0.25} & 99.51\textsubscript{$\pm$0.25} &99.75\textsubscript{$\pm$0.06} & 99.60\textsubscript{$\pm$0.09} \\
		& \multicolumn{2}{c||}{bin = 3 / 3.00} & 99.75\textsubscript{$\pm$0.16} & 99.76\textsubscript{$\pm$0.15} & 89.03\textsubscript{$\pm$20.58} & 92.39\textsubscript{$\pm$13.82} & 99.59\textsubscript{$\pm$0.17} & 99.52\textsubscript{$\pm$0.66} & 99.88\textsubscript{$\pm$0.10} & 99.80\textsubscript{$\pm$0.15} \\     
		\midrule[0.5pt]
		
		\multicolumn{1}{c}{\multirow{2}{*}{RNNstega \cite{RNN-Stega}}} & \multicolumn{2}{c||}{bit = 1 / 1.00} &    85.81\textsubscript{$\pm$1.93} & 86.03\textsubscript{$\pm$1.40} & 81.39\textsubscript{$\pm$1.70} & 82.59\textsubscript{$\pm$1.62} & 85.86\textsubscript{$\pm$0.50} & 86.29\textsubscript{$\pm$0.89} &97.42\textsubscript{$\pm$0.44} & 95.97\textsubscript{$\pm$0.67} \\
		& \multicolumn{2}{c||}{bit = 3 / 2.63} & 82.93\textsubscript{$\pm$2.18} & 84.48\textsubscript{$\pm$2.20} & 80.35\textsubscript{$\pm$2.38} & 77.94\textsubscript{$\pm$3.04} & 82.71\textsubscript{$\pm$0.17} & 83.10\textsubscript{$\pm$1.02} &96.78\textsubscript{$\pm$0.82} & 94.74\textsubscript{$\pm$1.34} \\
		\midrule[0.5pt]
		
		\multirow{1}{*}{ADG \cite{ADG}} & \multicolumn{2}{c||}{$\tau$ = 0.5 / 4.38}&  85.86\textsubscript{$\pm$1.73} & 86.35\textsubscript{$\pm$1.82} & 82.23\textsubscript{$\pm$2.40} & 82.08\textsubscript{$\pm$2.96} & 84.86\textsubscript{$\pm$1.74} & 84.10\textsubscript{$\pm$1.80} & 98.34\textsubscript{$\pm$0.39} & 97.41\textsubscript{$\pm$0.58} \\
		\midrule[0.5pt]
		
		LSCS \cite{Lu2023IJCNN}&\multicolumn{2}{c||}{- / 1.12}& 78.86\textsubscript{$\pm$3.07} &78.98\textsubscript{$\pm$2.83} & 81.19\textsubscript{$\pm$6.62}& 80.11\textsubscript{$\pm$8.26}&84.15\textsubscript{$\pm$0.41} & 85.11\textsubscript{$\pm$0.79}& 96.22\textsubscript{$\pm$2.08} &95.48\textsubscript{$\pm$3.11} \\
		\midrule[0.5pt]
		
		PLMmark \cite{Liu2023AAAI}&\multicolumn{2}{c||}{- / -}&	98.64\textsubscript{$\pm$0.92}&	98.60\textsubscript{$\pm$0.95}&	98.81\textsubscript{$\pm$1.83}&	98.75\textsubscript{$\pm$1.93}&	99.92\textsubscript{$\pm$0.17}&	99.92\textsubscript{$\pm$0.11}&	99.85\textsubscript{$\pm$0.30}&	99.85\textsubscript{$\pm$0.30}  \\
		\midrule[0.5pt]
		
		\multicolumn{1}{c}{\multirow{6}{*}{\textbf{DAIRstega}}} &\multirow{3}{*}{$\beta$ = 1}&$\alpha$ = 8 / 1.10&  59.50\textsubscript{$\pm$2.77} & 60.15\textsubscript{$\pm$10.59} & 56.82\textsubscript{$\pm$4.09} & 64.21\textsubscript{$\pm$10.03} & 67.25\textsubscript{$\pm$3.23} & 68.04\textsubscript{$\pm$1.53} & 75.40\textsubscript{$\pm$0.37} & 76.26\textsubscript{$\pm$1.28} \\
		&&$\alpha$ = 32 / 1.11& \underline{59.35\textsubscript{$\pm$1.25}*} & \underline{59.99\textsubscript{$\pm$5.51}*} & \textbf{54.05\textsubscript{$\pm$4.72}} & 61.68\textsubscript{$\pm$11.17} & \textbf{62.28\textsubscript{$\pm$0.50}} & \textbf{62.66\textsubscript{$\pm$2.31}} & 71.97\textsubscript{$\pm$0.64} & 72.75\textsubscript{$\pm$1.21} \\
		&&$\alpha$ = 48 / 1.13 & \textbf{58.00\textsubscript{$\pm$2.37}} & \textbf{56.08\textsubscript{$\pm$13.79}} & \underline{54.98\textsubscript{$\pm$3.14}*} & \textbf{59.17\textsubscript{$\pm$11.37}} & \underline{65.51\textsubscript{$\pm$2.23}*} & 66.78\textsubscript{$\pm$3.91} &  \underline{71.40\textsubscript{$\pm$0.65}*} & \underline{70.89\textsubscript{$\pm$2.05}*} \\
		\cdashline{2-11}[3pt/2pt]
		
		&\multirow{3}{*}{$\beta$ = 0.5}&$\alpha$ = 8 / 1.89&62.88\textsubscript{$\pm$1.53} &62.21\textsubscript{$\pm$1.83}	&63.13\textsubscript{$\pm$1.44}	&\underline{59.77\textsubscript{$\pm$5.36}*}	&65.67\textsubscript{$\pm$1.56}	&\underline{65.84\textsubscript{$\pm$2.20}*}	&\textbf{69.04\textsubscript{$\pm$0.36}}	&\textbf{68.23\textsubscript{$\pm$2.50}}\\
		&&$\alpha$ = 32 / 2.56&72.06\textsubscript{$\pm$1.17}	&71.78\textsubscript{$\pm$1.21}	&68.73\textsubscript{$\pm$1.29}	&68.19\textsubscript{$\pm$3.44}	&71.71\textsubscript{$\pm$1.95}	&70.34\textsubscript{$\pm$2.41}	&76.41\textsubscript{$\pm$2.56}	&75.90\textsubscript{$\pm$2.73}\\
		&&$\alpha$ = 48 / 2.58&70.17\textsubscript{$\pm$1.81}	&70.40\textsubscript{$\pm$3.11}	&69.48\textsubscript{$\pm$0.68}	&67.80\textsubscript{$\pm$1.98}	&70.64\textsubscript{$\pm$1.61}	&71.00\textsubscript{$\pm$2.23}	&79.46\textsubscript{$\pm$0.56}	&78.28\textsubscript{$\pm$1.13}\\
		\bottomrule[1.2pt]
	\end{tabular}%
\end{table*}%

\subsection{Supplement data for anti-steganalysis}\label{Appendix_D}
We also perform other 7 steganalysis methods (including non-BERT-based, BERT-based, and LLM-based steganalysis) to detect the stegos we generated, as shown in Table \ref{llm-stega}, where, \cite{tmode} is the current method that has achieved SOTA (State-of-the-art) steganalysis performances.
\begin{table}[H]
	\centering
	\small
	\setlength{\tabcolsep}{0.51mm}
	\caption{The supplemental anti-steganalysis capability of the DAIRstega from other 7 steganalysis works. \textbf{Bold} represents the best specific and Avg. result. ``\underline{ *}" represents the suboptimal specific result.}\label{llm-stega}%
	\begin{tabular}{c|cc||cccccc|c||cccccc|c}
		\toprule[1.2pt]
		\multicolumn{3}{c||}{\multirow{2.2}*{\textbf{Supplemental}}} & \multicolumn{7}{c||}{VAE-Stega \cite{VAE-Stega}}                & \multicolumn{7}{c}{\textbf{DAIRstega}} \\
		\cdashline{4-17}[2pt/2pt]
		
		\multicolumn{3}{c||}{\multirow{2.1}*{\textbf{anti-steganalysis}}} & \multicolumn{3}{c}{AC} & \multicolumn{3}{c|}{HC} &\multirow{2}{*}{Avg.}& \multicolumn{2}{c}{$\alpha$ = 8} & \multicolumn{2}{c}{$\alpha$ = 16} & \multicolumn{2}{c|}{$\alpha$ = 32}&\multirow{2}{*}{Avg.}\\
		\cdashline{4-9}[2pt/2pt]\cdashline{11-16}[2pt/2pt]   \multicolumn{3}{c||}{} & Movie & Twitter & News & Movie & Twitter & News & & $\beta$ = 1   & $\beta$ = 0.5 & $\beta$ = 1   & $\beta$ = 0.5 & $\beta$ = 1   & $\beta$ = 0.5& \\
		\midrule[0.5pt]
		\multirow{3}{*}{non-BERT-}&\multirow{2}{*}{\cite{FEFT}} & Acc $\downarrow$   & 57.63  & 53.75  & 52.55  & 61.63  & 57.75  & 74.38  & 59.61 & 54.38 & \textbf{51.37}  & \underline{51.60*}  & 56.97  & 52.35  & 59.50 &\textbf{54.38}  \\
		&& F1 $\downarrow$    & 50.65  & \textbf{34.63}  & \underline{46.31*}  & 60.89  & 66.65  & 73.89  &55.50 & 47.09  & 47.55  & 49.47  & 52.35  & 49.20  & 57.70&\textbf{50.56} \\
		\cline{2-17}
		
		based&\multirow{2}{*}{\cite{rbilstmc}} & Acc $\downarrow$   & 64.50  & \textbf{58.75}  & 62.50  & 87.63  & 80.75  & 94.38  &74.75 
		& 64.75  & \underline{61.05*}  & 61.88  & 67.42  & 63.83  & 68.73&  \textbf{64.61} \\
		&& F1 $\downarrow$    & 66.91  & \textbf{56.46}  & 64.95  & 88.03  & 82.14  & 94.33  &75.47 
		& 63.85  & \underline{61.05*}  & 67.03  & 66.16  & 63.82  & 68.83&  \textbf{65.12} \\
		\midrule[0.5pt]
		
		\multirow{5}{*}{BERT-}&\multirow{2}{*}{\cite{Zou}} & Acc $\downarrow$   & 87.38  & 74.03  & 92.25  & 92.75  & 86.81  & 97.38  &88.43 
		& 70.63  & \underline{65.43*}  & \textbf{65.06}  & 65.87  & 65.56  & 73.25& \textbf{67.63}  \\
		&& F1 $\downarrow$    & 87.39  & 74.21  & 91.99  & 93.21  & 86.85  & 97.35  &88.50 
		& 70.44  & \textbf{61.48}   & 65.28  & \underline{65.15*} & 67.53  & 74.27& \textbf{67.36} 	 \\
		\cline{2-17}
		
		&\multirow{2}{*}{\cite{SSLS}} & Acc $\downarrow$   & 90.75  & 78.75  & 95.25  & 95.25  & 88.38  & 98.13  &91.08 
		& 70.38  & \underline{65.50*}  & \textbf{64.38}  & 71.12  & \underline{65.50*}  & 74.25& \textbf{68.52}  \\
		based&& F1 $\downarrow$    & 90.74  & 78.75  & 95.25  & 95.25  & 88.36  & 98.12 & 91.08 & 70.34  & 65.50  & \textbf{64.27}  & 70.99  & \underline{65.31*}  & 74.25& \textbf{68.44}	\\
		\cline{2-17}
		
		&\multirow{2}{*}{\cite{sesy}} & Acc $\downarrow$   & 92.50  & 75.38  & 95.88  & 94.75  & 88.32  & 97.62  &90.74
		& 70.12  & \textbf{64.25}  & \textbf{64.25}  & 73.75  & \underline{66.50*}  & 73.88& \textbf{68.79}  \\
		&& F1 $\downarrow$    & 92.19  & 73.20  & 95.62  & 94.53  & 88.26  & 97.60  & 90.23
		& 72.31  & \textbf{66.19}  & \underline{66.29*}  & 71.47  & 70.02  & 69.04&\textbf{69.22}  \\
		\midrule[0.5pt]
		
		\multirow{3}{*}{LLM-}&\multirow{2}{*}{\cite{llmsteganalysis}} & Acc $\downarrow$   & 89.63  & 78.13  & 93.13  & 88.63  & 76.38  & 86.75  & 85.44 
		& 63.38  &\underline{59.63*}  & 61.44  & 63.25  & \textbf{59.38}  & 62.13 & \textbf{61.53}\\
		&& F1 $\downarrow$    & 89.59  & 78.06  & 93.12  & 88.62  & 76.13  & 86.75 &85.38 
		& 63.24  & \textbf{60.15}  & \underline{60.59*}  & 63.13  & 59.15  & 61.86& \textbf{61.35} \\
		\cline{2-17}

		based&\multirow{2}{*}{\cite{tmode}} & Acc $\downarrow$   & 97.38 & 92.38  & 98.38  & 98.13  & 93.88  & 98.88  &96.50 
		& 75.38 & 72.63  & \underline{71.75*}  & 81.75  &\textbf{69.38}  & 83.25 &\textbf{75.69}  \\
		&& F1 $\downarrow$    & 96.87  & 92.97  & 98.57  & 97.10  & 92.97  & 98.97  &96.25 
		& 75.27  & 72.41  & \underline{71.35*}  & 81.55  & \textbf{69.32}  & 83.15 & \textbf{75.51} \\
		\bottomrule[1.2pt]
	\end{tabular}%
\end{table}%

\subsection{Example of cross-modal secret information hiding}\label{Appendix_E}

\begin{table*}[!htbp]
	\centering
	\footnotesize
	\setlength{\tabcolsep}{0.7mm}
	\caption{Examples of longer stegos generated by DAIRstega using cross-modal information. Bitstream is the binary corresponding to the secret.}\label{cm}%
	\begin{tabular}{p{7em}p{8em}p{39em}cc}
		\toprule[1.2pt]
		\multicolumn{1}{p{6.5em}}{\footnotesize{Secret\&Bitsream}} & \multicolumn{1}{p{5em}}{\footnotesize{Instruction}} & \multicolumn{1}{c}{\footnotesize{Stegos}} & \multicolumn{1}{p{2em}}{\footnotesize{PPL}} & \multicolumn{1}{p{3em}}{\footnotesize{$\Delta$LDA}} \\
		\midrule[0.5pt]
		
		\multicolumn{1}{p{6.5em}}{\multirow{2}{=}{\newline{}\textbf{\footnotesize Secret}:\newline{} \begin{minipage}[b]{0.1\columnwidth}
					\centering
					\raisebox{-.5\height}{\includegraphics[width=\linewidth]{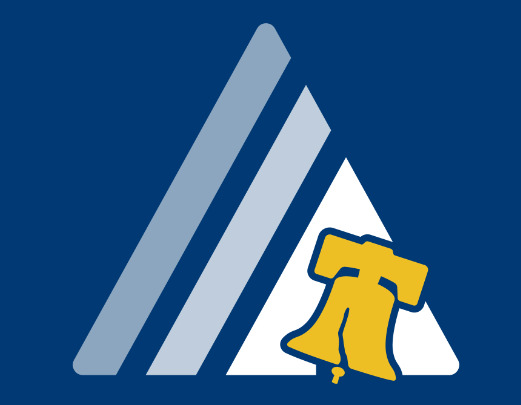}} 
				\end{minipage}
				\newline{}\newline{}\newline{} \textbf{\footnotesize{Bitstream}}:\newline{} 
				100010010101 000001001110 010001110000 110100001010 000110100000 101000000000 000000000000 000000001101 010010010100 100001000100 010100100000 000000000000 000000000110 110000000000 00...(31520 bit)}} & \multicolumn{1}{p{3.5em}}{\multirow{1}{=}{ \newline{}\newline{} Discuss the \newline{} causes of the \newline{} Great Depre-\newline{}ssion.}} & The Great Depression was a severe worldwide economic depression that took place from 1929 to 1939. It was the longest and most severe depression ever experienced by the industrialized world, surpassing in overall duration the Great Depression of 1873. The Depression began in the United States after the stock market crash of October 1929, and peaked between 1930 and 1933. By some measures, the Great Depression was the longest economic depression in the history of the industrialized world. The Great Depression had many causes, including overproduction, lack of consumer demand, high tariffs, banking crisis, stock market crash, deflation, wage reductions, and the 1929 collapse of Worldwide prices for agricultural commodities. The collapse of agricultural prices resulted from overproduction, due to improvements in seed stock, fertilizers, pesticides, transportation, and farm equipment, as well as increases in acreage and livestock numbers. & \multicolumn{1}{p{2em}}{\multirow{1}{=}{\newline{}\newline{}\newline{}\newline{}\newline{} 2.561}} & \multicolumn{1}{p{2em}}{\multirow{1}{=}{\newline{}\newline{}\newline{}\newline{}\newline{} 0.021}} \\
		\cline{2-5}          & \multicolumn{1}{p{3.5em}}{\multirow{1}{=}{\newline{}\newline{}\newline{}\newline{} Write a general overview of quantum computing.\newline{}\newline{}\newline{}\newline{}...}} & Quantum computing is a new and revolutionary form of computing that is based on the principles of quantum mechanics. Unlike traditional computing, which uses binary bits to represent information, quantum computing uses quantum bits (qubits) which can exist in a superposition of 0 and 1 simultaneously. This means that quantum computers can process information exponentially faster than traditional computers, and they can also solve certain types of problems that are intractable for traditional computers. Quantum computing has the potential to revolutionize many fields, including machine learning, cryptography, drug discovery, and weather forecasting. For example, quantum computers can be used to train machine learning models more efficiently, which could lead to more accurate and efficient models. They can also be used to break current encryption methods, which could have serious implications for cybersecurity. In the field of drug discovery, quantum computers can be used to simulate molecular interactions, which could lead to the discovery of new and more effective drugs. Finally, in the field of weather forecasting, quantum computers can be used to simulate the complex interactions between different weather systems, which could lead to more accurate and precise forecasts. \newline{} ...& \multicolumn{1}{p{2em}}{\multirow{1}{=}{\newline{}\newline{}\newline{}\newline{}\newline{}\newline{}\newline{} 1.915}}   & \multicolumn{1}{p{2em}}{\multirow{1}{=}{\newline{}\newline{}\newline{}\newline{}\newline{}\newline{}\newline{} 0.009}} \\
		
		\bottomrule[1.2pt]
	\end{tabular}%
\end{table*}%

\end{document}